\pdfoutput=1
\documentclass[11pt]{article}

\usepackage[]{EMNLP2023}

\usepackage{times}
\usepackage{latexsym}

\usepackage[T1]{fontenc}
\usepackage[utf8]{inputenc}

\usepackage{microtype}

\usepackage{inconsolata}

\usepackage{amsmath}

\usepackage{multirow}
\usepackage{multicol}
\usepackage{makecell}
\usepackage{graphicx} 
\usepackage{amssymb}
\usepackage{algorithm}
\usepackage{algpseudocode}
\usepackage{ulem}
\usepackage{booktabs}
\usepackage{CJKutf8}
\usepackage{subfigure}

\usepackage{arydshln}
\usepackage{enumitem}

\title{Learning Multilingual Sentence Representations with Cross-lingual Consistency Regularization}

\author{Pengzhi Gao, Liwen Zhang, Zhongjun He, Hua Wu, and Haifeng Wang \\
Baidu Inc. No. 10, Shangdi 10th Street, Beijing, 100085, China \\
\texttt{\{gaopengzhi,zhangliwen04,hezhongjun,wu\_hua,wanghaifeng\}@baidu.com} 
}

\begin{document}
\maketitle

\begin{abstract}

Multilingual sentence representations are the foundation for similarity-based bitext mining, which is crucial for scaling multilingual neural machine translation (NMT) system to more languages. In this paper, we introduce MuSR: a one-for-all \textbf{Mu}ltilingual \textbf{S}entence \textbf{R}epresentation model that supports more than 220 languages. Leveraging billions of English-centric parallel corpora, we train a multilingual Transformer encoder, coupled with an auxiliary Transformer decoder, by adopting a multilingual NMT framework with CrossConST, a cross-lingual consistency regularization technique proposed in \citet{gao2023crossconst}. Experimental results on multilingual similarity search and bitext mining tasks show the effectiveness of our approach. Specifically, MuSR achieves superior performance over LASER3\footnote{In its original context, LASER3 refers solely to the language-specific models presented in \citet{heffernan-etal-2022-bitext}. For simplicity, we use LASER3 as an umbrella term encompassing the multilingual model LASER2 and the language-specific models discussed in this paper.} \cite{heffernan-etal-2022-bitext} which consists of 148 independent multilingual sentence encoders. % Our implementation and the pretrained models are available at \url{https://github.com/gpengzhi/CrossConST-SR}.

\end{abstract}

\section{Introduction}

Multilingual sentence representation models \cite{artetxe-schwenk-2019-massively,yang-etal-2020-multilingual,reimers-gurevych-2020-making,feng-etal-2022-language,heffernan-etal-2022-bitext,mao-nakagawa-2023-lealla} align different languages in a shared representation space, facilitating similarity-based bitext mining that extracts parallel sentences for learning multilingual neural machine translation (NMT) systems \cite{schwenk-etal-2021-wikimatrix,schwenk-etal-2021-ccmatrix}. Specifically, LASER3 \cite{heffernan-etal-2022-bitext} scales the original LASER \cite{artetxe-schwenk-2019-massively} beyond the 93 widely used languages and achieves the state-of-the-art (SOTA) performance on the multilingual sentence alignment tasks over 200 languages.

\begin{figure}[h]
\centering
\includegraphics[scale=0.65]{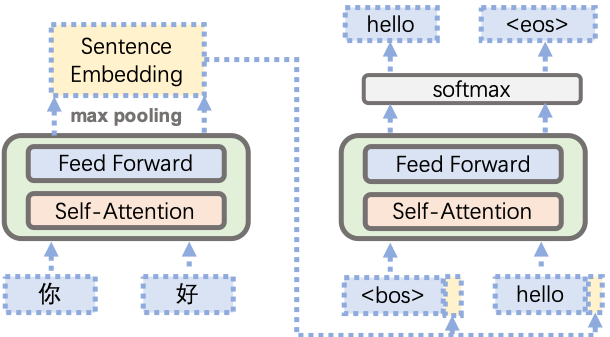}
\caption{The model architecture of our approach for learning multilingual sentence representations.}
\label{fig:arch}
\end{figure}

Although LASER3 exhibits remarkable performance, it is not a one-for-all multilingual sentence representation model. Instead, it comprises of one multilingual model called LASER2 and 147 language-specific models, which are learned through a teacher-student training mechanism. Such model strategy, although effective, results in substantial storage overhead of 78GB and degraded transfer performance from high-resource to low-resource languages, which hinders its practical value in natural language processing (NLP).

\begin{table*}\small
\centering
\begin{tabular}{c|c|c|c|c|c|c|c}
Method & \#Models & \#Parameters & \#Languages & Task & Architecture & Monolingual & Pretrain \\
\hline
\hline
LASER2 & 1 & 45M & 93 & Seq2Seq & Bi-LSTM & & \\
LASER3 & 1 + 147 & N/A & 205 & Dual Encoder & Transformer & $\checkmark$ \\
% LEALLA & 1 & 69/107/147 M & 109 & Dual Encoder & Transformer & & \\
LaBSE & 1 & 471M & 109 & Dual Encoder & Transformer & $\checkmark$ & $\checkmark$ \\
\hline
MuSR & 1 & 434M & 223 & Seq2Seq & Transformer & & \\
\end{tabular}
\caption{Comparison between the related works and our approach. Note that language-specific models in LASER3 have different vocabulary size, and the number of parameters for each model can be approximately calculated as $202\text{M} + \text{vocabulary size} \times 1024$. ``Monolingual'' denotes whether the monolingual data is used for training. ``Pretrain'' denotes whether the model relies on the language model pretraining.}
\label{tab:comparison}
\end{table*}

In this paper, our primary goal is to learn a unified multilingual sentence encoder, MuSR, to handle a wide range of languages such that semantic-equivalent sentences in different languages are close to each other in the representation space. Inspired by the cross-lingual consistency for multilingual NMT \cite{gao2023crossconst}, we learn multilingual sentence embeddings by utilizing a many-to-one multilingual NMT training paradigm with cross-lingual consistency regularization (Figures \ref{fig:arch} and \ref{fig:strategy}). In order to support a wide range of languages, we collect about $5.5$ billion English-centric parallel sentences covering more than 220 languages from both open-source and in-house datasets. To the best of our knowledge, MuSR is the first one-for-all multilingual sentence representation model that supports more than 220 languages. The contributions of this paper can be summarized as follows:

\begin{itemize}
\item We learn a one-for-all multilingual sentence representation model, MuSR, by leveraging many-to-one multilingual NMT training with CrossConST regularization over $5.5$ billion English-centric parallel corpora.
\item Our experimental results show that MuSR achieves impressive performance on the multilingual benchmarks and outperforms the SOTA models LaBSE \cite{feng-etal-2022-language} and LASER3 \cite{heffernan-etal-2022-bitext}.
\item We publicly release MuSR, the multilingual sentence representation model that supports 223 languages.\footnote{Our implementations are available at \url{https://github.com/gpengzhi/CrossConST-SR}.}
\end{itemize}

\section{Background}

\subsection{Multilingual Sentence Representation}

As an important component of cross-lingual and multilingual NLP, multilingual sentence representation has attracted increasing attention in the NLP community. One direction is to leverage dual-encoder architecture to learn language-agnostic representations. \citet{guo-etal-2018-effective} demonstrate the effectiveness of the dual-encoder model for learning bilingual sentence embeddings, and \citet{ijcai2019p746} extend the dual-encoder model with additive margin softmax loss. Based on these works, LaBSE \cite{feng-etal-2022-language} utilizes dual Transformer encoders to learn language-agnostic embeddings over 109 languages with additive margin softmax loss, which is also pretrained with masked language modeling (MLM) and translation language modeling (TLM) \cite{NEURIPS2019_c04c19c2}. LEALLA \cite{mao-nakagawa-2023-lealla} further constructs low-dimensional sentence embeddings by leveraging knowledge distillation based on LaBSE.

Another direction is to utilize encoders from multilingual NMT to produce universal representations across different languages. LASER \cite{artetxe-schwenk-2019-massively} learns the multilingual sentence embeddings over 93 languages based on the NMT model with a Bi-LSTM encoder and a LSTM decoder. \citet{heffernan-etal-2022-bitext} replace the original LASER model with LASER2 by introducing SentencePiece \cite{kudo-richardson-2018-sentencepiece} vocabulary, up-sampling the low-resource languages, and adopting a new fairseq\footnote{\url{https://github.com/facebookresearch/fairseq}} implementation. LASER2 is used as the teacher, and 147 language-specific sentence representation models are learned by utilizing teacher-student and MLM training mechanisms. LASER3 refers to a group of LASER2 and 147 language-specific models across 205 languages. The comparison between the existing works and our approach are summarized in Table \ref{tab:comparison}.

\subsection{Cross-lingual Consistency Regularization for Multilingual NMT}
 
The multilingual NMT model refers to a neural network with an encoder-decoder architecture, which receives a sentence in language $L_i$ as input and returns a translated sentence in language $L_j$ as output. Assume $\mathbf{x}$ and $\mathbf{y}$ correspond to the source and target sentences respectively, and let $\mathcal{S}$ denotes the multilingual training corpus. The standard training objective is to minimize the empirical risk:
\begin{equation}
\mathcal{L}_{ce}(\theta) =  \mathop{\mathbb{E}}\limits_{(\mathbf{x}, \mathbf{y}) \in \mathcal{S}} [\ell(f(\mathbf{x}, \mathbf{y}; \theta), \ddot{\mathbf{y}})],
\end{equation}
where $\ell$ denotes the cross-entropy loss, $\theta$ is a set of model parameters, $f(\mathbf{x}, \mathbf{y}; \theta)$ is a sequence of probability predictions, i.e., 
\begin{equation}
f_j(\mathbf{x}, \mathbf{y}; \theta) = P(y|\mathbf{x}, \mathbf{y}_{<j}; \theta),
\end{equation}
and $\ddot{\mathbf{y}}$ is a sequence of one-hot label vectors for $\mathbf{y}$.

\begin{figure*}[h]
\centering
\includegraphics[scale=0.5]{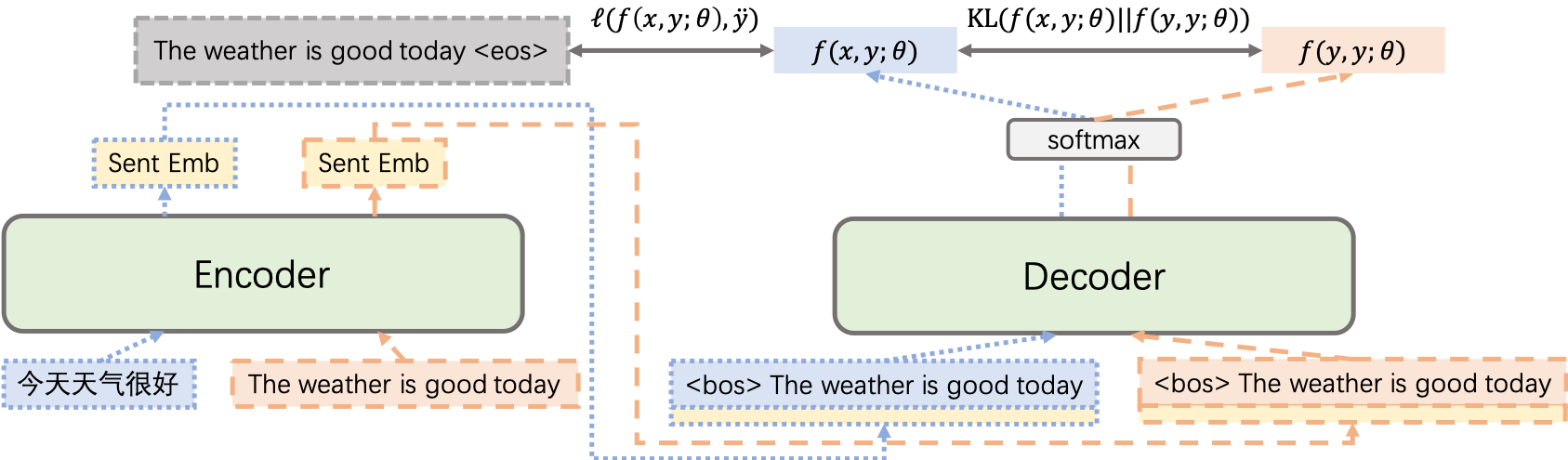}
\caption{Illustration of CrossConST regularization for learning multilingual sentence representations, where the original Chinese-English sentence pair ("\begin{CJK}{UTF8}{gbsn}今天天气很好\end{CJK}", "The weather is good today") and the copied English-English sentence pair ("The weather is good today", "The weather is good today") are fed into the multilingual NMT model to generate two output distributions $f(\mathbf{x}, \mathbf{y}; \theta)$ and $f(\mathbf{y}, \mathbf{y}; \theta)$.}
\label{fig:strategy}
\end{figure*}

\citet{gao2023crossconst} introduce a cross-lingual consistency regularization, CrossConST, to bridge the representation gap among different languages in the training of multilingual NMT model. For each sentence pair $(\mathbf{x}, \mathbf{y})$, the training objective of CrossConST is defined as:
\begin{equation}\label{main_loss}
\mathcal{L}_{CrossConST}(\theta) = \mathcal{L}_{ce}(\theta) + \alpha \mathcal{L}_{kl}(\theta),
\end{equation}
where
\begin{equation}\label{kl_constraint}
\mathcal{L}_{kl}(\theta) = \text{KL}(f(\mathbf{x}, \mathbf{y}; \theta) \| f(\mathbf{y}, \mathbf{y}; \theta)),
\end{equation}
$\text{KL}(\cdot \| \cdot)$ denotes the Kullback-Leibler (KL) divergence between two distributions, and $\alpha$ is a scalar hyper-parameter that balances $\mathcal{L}_{ce}(\theta)$ and $\mathcal{L}_{kl}(\theta)$.

\section{Methodology}\label{sec:methodology}

% e adopt a single multilingual Transformer encoder for generating language agnostic sentence representations. 
Following the similar problem formulation of \citet{artetxe-schwenk-2019-massively}, our approach is based on a Transformer encoder-decoder architecture trained with English-centric parallel corpora. We discuss the details of our model architecture and training strategy as follows.

\subsection{Model Architecture}\label{sec:architecture}

The overall model architecture is illustrated in Figure \ref{fig:arch}. Multilingual sentence embeddings are calculated by applying a max-pooling operation over the Transformer encoder's output, which is subsequently concatenated to the word embeddings at the Transformer decoder's input. We discard the cross-attention module in the Transformer decoder, and the sentence embeddings are the only connection between the encoder and the decoder such that all relevant information of the input sentences are captured by the corresponding sentence representations. Note that our model does not need language tags, as many-to-one multilingual NMT does not rely on them, unlike LASER in \citet{artetxe-schwenk-2019-massively}.

\subsection{Training Strategy}\label{sec:training_strategy}

Following \citet{gao2023crossconst}, we adopt a two-stage training strategy to stabilize the multilingual NMT training procedure and accelerate the convergence of the multilingual NMT model. Instead of utilizing two target languages (English and Spanish) as in \citet{artetxe-schwenk-2019-massively}, we consider only one target language (English) and formulate our problem as a many-to-one multilingual NMT task. We first train a multilingual NMT model as the pretrained model and then finetune the model with CrossConST objective function \eqref{main_loss}. Figure \ref{fig:strategy} illustrates CrossConST regularization for learning multilingual sentence representations. Through the application of CrossConST, sentence embeddings of the target language are aligned to the representation space of the source languages. The alignment process is facilitated by our many-to-one multilingual NMT model, which effectively encodes all languages into a shared representation space.

\section{Datasets and Training Configurations}

\subsection{Datasets}

We use a combination of open-source datasets and in-house datasets in our experiments.\footnote{See the list of the supported languages in Table \ref{tab:langs}.}

\begin{figure*}[h]
\centering
\includegraphics[scale=0.7]{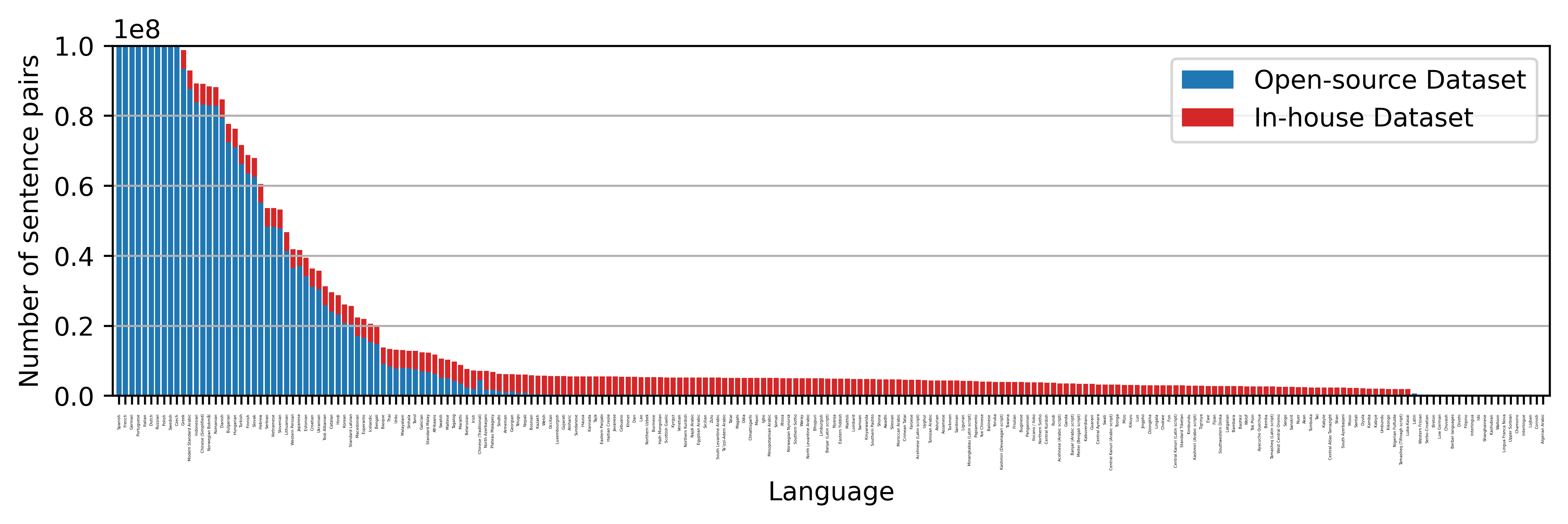}
\caption{The distribution of the open-source and in-house cleaned datasets for each language in our training dataset. Note that the sentences for each language are capped at 100 million for better illustration. Please check Figure \ref{fig:distribution_full} for the complete distribution.}
\label{fig:distribution}
\end{figure*}

\paragraph{Open-source Dataset} We collect all English-centric parallel datasets from the OPUS collection\footnote{\url{http://www.opus.nlpl.eu}} \cite{TIEDEMANN12.463} up to October 2022, which is comprised of multiple corpora, ranging from movie subtitles \cite{tiedemann-2016-finding} to Bible \cite{christodouloupoulos2015massively} to web crawled datasets \cite{el-kishky-etal-2020-ccaligned,schwenk-etal-2021-ccmatrix}. We download all available English-centric corpora and concatenate them without curating the datasets or trying to balance the representation of different domains.

\paragraph{In-house Dataset} We also leverage all English-centric in-house datasets which consists of the following resources: 1) The parallel sentences are constructed from web pages by utilizing a bitext mining system. The extracted sentence pairs are filtered by a predefined scoring threshold. 2) We adopt the 3.3B multilingual NMT model released by the No Language Left Behind (NLLB) project\footnote{\url{https://github.com/facebookresearch/fairseq/tree/nllb}} and translate the English sentences from the ParaCrawl project\footnote{\url{https://opus.nlpl.eu/ParaCrawl.php}} \cite{banon-etal-2020-paracrawl} into different languages. 3) We leverage our in-house multilingual NMT model to translate the in-house English corpus into different languages.

After we collect all parallel datasets, we adopt the data cleaning process as follows: 1) We remove duplicate sentence pairs and also discard sentence pairs wherein the English sentences exceed 5000 characters. 2) Language identification filtering is applied by utilizing fastText toolkit \cite{joulin2016fasttext,joulin-etal-2017-bag}. If the language is not supported by the identification model\footnote{\url{https://fasttext.cc/docs/en/language-identification.html}}, we simply check whether the language is non-English. 3) Dual conditional cross-entropy filtering \cite{junczys-dowmunt-2018-dual} is performed based on our in-house multilingual NMT models. Specifically, for a sentence pair $(\mathbf{x}, \mathbf{y})$, we identify they are translations of each other by leveraging the score defined as follows:
\begin{equation}\nonumber
|H(y|x) - H(x|y)| + \frac{1}{2}(H(y|x) + H(x|y)),
\end{equation}
where $H(\cdot|\cdot)$ denotes the word-normalized conditional cross-entropy loss based on the multilingual NMT model. After the cleaning process, we discard the languages which have less than 1000 sentence pairs. In summary, we collect about 5.5 billion cleaned English-centric sentence pairs covering 223 languages including English. The distribution of our training datasets for each language is illustrated in Figure \ref{fig:distribution}.

% \paragraph{Distilled from MT model} Additionally, we employed the NLLB 3.3B model and our in-house model to translate English sentences from the Paracrawl \cite{banon-etal-2020-paracrawl} dataset, yielding around 5 million manually translated sentence pairs across over 190 language directions.

% Ultimately, we acquired 232 language directions with approximately 5.2 billion sentence pairs. The detailed data distribution can be found in Appendix A.

% We employed basic data cleaning procedures, such as removing duplicate sentences and utilizing fastText \cite{joulin2016bag} for language identification within each language. For languages not covered by FastText's 176 languages, we simply determined if they were not English. We retained language pairs with a remaining corpus of over 10,000 sentence pairs, leading to a total of 140 languages and approximately 4.6 billion sentence pairs. 

% The distribution of the training dataset for each language is shown in Figure \ref{fig:distribution}.

We can see that there is a discrepancy of 5 orders of magnitude between the highest (Spanish) and the lowest (Algerian Arabic) resource languages. To strike a balance between high and low resource language pairs, we adopt a temperature-based sampling strategy \cite{arivazhagan2019massively, bapna-firat-2019-simple}. Sentence pairs are sampled according to a multinomial distribution with probability $\{q_i\}_{i=1,...,N}$, where
\begin{equation}
q_i = \frac{p^{\alpha}_i}{\sum_{j=1}^Np^{\alpha}_j} \quad \text{with} \quad  p_i = \frac{n_i}{\sum_{k=1}^Nn_k},
\end{equation}
$N$ denotes the number of languages, and $n_i$ denotes the number of sentence pairs for each language. We consider $\alpha=0.5$ in our experiments. Sampling with this distribution increases the number of sentence pairs associated to low resource languages and alleviates the bias towards high resource languages. We collect 500 million sentences with such sampling strategy and learn a shared dictionary with 256K byte-pair-encoding (BPE) \cite{sennrich-etal-2016-neural} types using SentencePiece\footnote{\url{https://github.com/google/sentencepiece}}. We keep tokens occurring no less than $20$, which results in a subword vocabulary of $344,276$ tokens.

% learn a joint SentencePiece \cite{kudo-richardson-2018-sentencepiece} model with 256K tokens. 

% Let $D_i$ denote the amount of training data for language pair $L_i$. If we sample from the union of the datasets, the probability of a sample belonging to language pair $p_i = \frac{D_i}{\sum_{M}^j D_j}$, where M represents the number of supported languages. We set the probability of our sampled distribution to be proportional to $p_i^{\frac{1}{T}}$, where $T$ is the sampling temperature. Here, $T = 1$ corresponds to the true data distribution, and $T = 100$ results in an (almost) equal number of samples for each language. We used $T = 5$ for our experiments.

\subsection{Training Configurations}

We implement our approach on top of the Transformer \cite{vaswani2017attention}. We apply a Transformer with $12$ encoder layers and $3$ decoder layers, $8$ attention heads, embedding size $768$, and FFN layer dimension $768 \times 4$ and $768 \times 2 \times 4$ for encoder and decoder respectively. We apply cross-entropy loss with label smoothing rate $0.1$ and set max tokens per batch to be $1024$. We use the Adam optimizer with Beta $(0.9, 0.98)$, $10000$ warmup updates, and inverse square root learning rate scheduler with initial learning rates $7e^{-4}$. We set max source positions and max target positions to be $256$ and use dropout rate $0.1$. We apply the same training configurations in both pretraining and finetuning stages. We fix $\alpha$ to be $1.0$ in \eqref{main_loss} for CrossConST. We train all models until convergence on $8 \times 4$ NVIDIA Tesla V100 GPUs.

\section{Experimental Evaluation}

% In this paper, we focus on the application of multilingual sentence representations on similarity-based bitext mining for boosting NMT performance. Note that NMT model training is computationally expensive, and it is intractable to evaluate the performance of various models. 
Following the evaluation setup of \citet{heffernan-etal-2022-bitext}, we here investigate the performance of multilingual sentence embeddings on two benchmarks: multilingual similarity search and bitext mining.

\subsection{Multilingual Similarity Search}

Given the parallel sentence pairs, we find the nearest neighbor for each sentence in the other language according to the sentence embedding cosine similarity and compute the corresponding accuracy. We conduct our experiments on the following datasets:

\paragraph{Tatoeba} Tatoeba is a multilingual dataset covering 112 languages \cite{artetxe-schwenk-2019-massively}, which contains up to $1000$ sentences per language along with their English translations.\footnote{\url{https://github.com/facebookresearch/LASER/tree/main/data/tatoeba/v1}}

\paragraph{Flores-200} Flores-200 is a multilingual dataset made publicly available by the NLLB project \cite{costa2022no}, which covers 204 languages.%, and consists of translations from 842 distinct web articles, totaling 3001 sentences for each language.
\footnote{\url{https://github.com/facebookresearch/flores/tree/main/flores200}} %The sentences are divided into three splits: dev, devtest, and test, and 
We perform the evaluation on the devtest which includes 1012 sentences for each language. We also evaluate on Flores-101 which is a subset of Flores-200 and covers $102$ languages.

\begin{table}[h]\small
\centering
\begin{tabular}{c | c c c c c} 
\multicolumn{1}{c|}{Model} & \multicolumn{1}{c}{Tatoeba} & \multicolumn{2}{c}{Flores-101} & \multicolumn{2}{c}{Flores-200} \\
& $\leftrightarrow$ \texttt{en} & $\leftrightarrow$ \texttt{en} & $\leftrightarrow$ \texttt{zh} & $\leftrightarrow$ \texttt{en} &  $\leftrightarrow$ \texttt{zh} \\
\hline
\hline
LASER2 & 69.95 & 67.78 & 64.47 & 56.98 & 52.76 \\
% LEALLA & 83.39 & 95.83 & 94.64 & 87.90 & 85.41 \\
LaBSE & 83.23 & 96.43 & 95.46 & 88.48 & 86.06 \\
LASER3 & 78.08 & 98.30 & 96.18 & 93.71 & 90.64 \\
\hline
MuSR & \bf 83.96 & \bf 99.23 & \bf 98.48 & \bf 97.37 & \bf 95.95 \\
\end{tabular}
\caption{Our approach achieves the superior performance over the existing SOTA models on the Tatoeba and Flores benchmarks. The detailed experimental results are summarized in Tables \ref{tab:tatoeba}, \ref{tab:flores200-en-1}, \ref{tab:flores200-en-2}, \ref{tab:flores200-zh-1}, and \ref{tab:flores200-zh-2}.}
\label{mss_result}
\end{table}

We report the averaged bidirectional similarity search accuracy with English (\texttt{en}) and Chinese (\texttt{zh}) on the Tatoeba, Flores-101, and Flores-200 benchmarks in Table \ref{mss_result}. The English direction represents the supervised performance of MuSR, while the Chinese direction exemplifies the effectiveness in the zero-shot scenario. We can see that our approach significantly outperforms the current SOTA models LaBSE and LASER3. It is worth mentioning that MuSR achieves an improvement of over $3.5\%$ accuracy on average over LASER3 that consists of $148$ independent sentence embedding models. The performance gap between English and Chinese in LaBSE, the model with the smallest discrepancy, stands at $0.97\%$ and $2.42\%$ on Flores-101 and Flores-200 respectively. In contrast, MuSR exhibits a substantially smaller divergence of $0.75\%$ and $1.42\%$ on these two directions, indicating our superior capability to model various languages within the shared representation space.

\begin{figure}[h]
\centering
\includegraphics[scale=0.35]{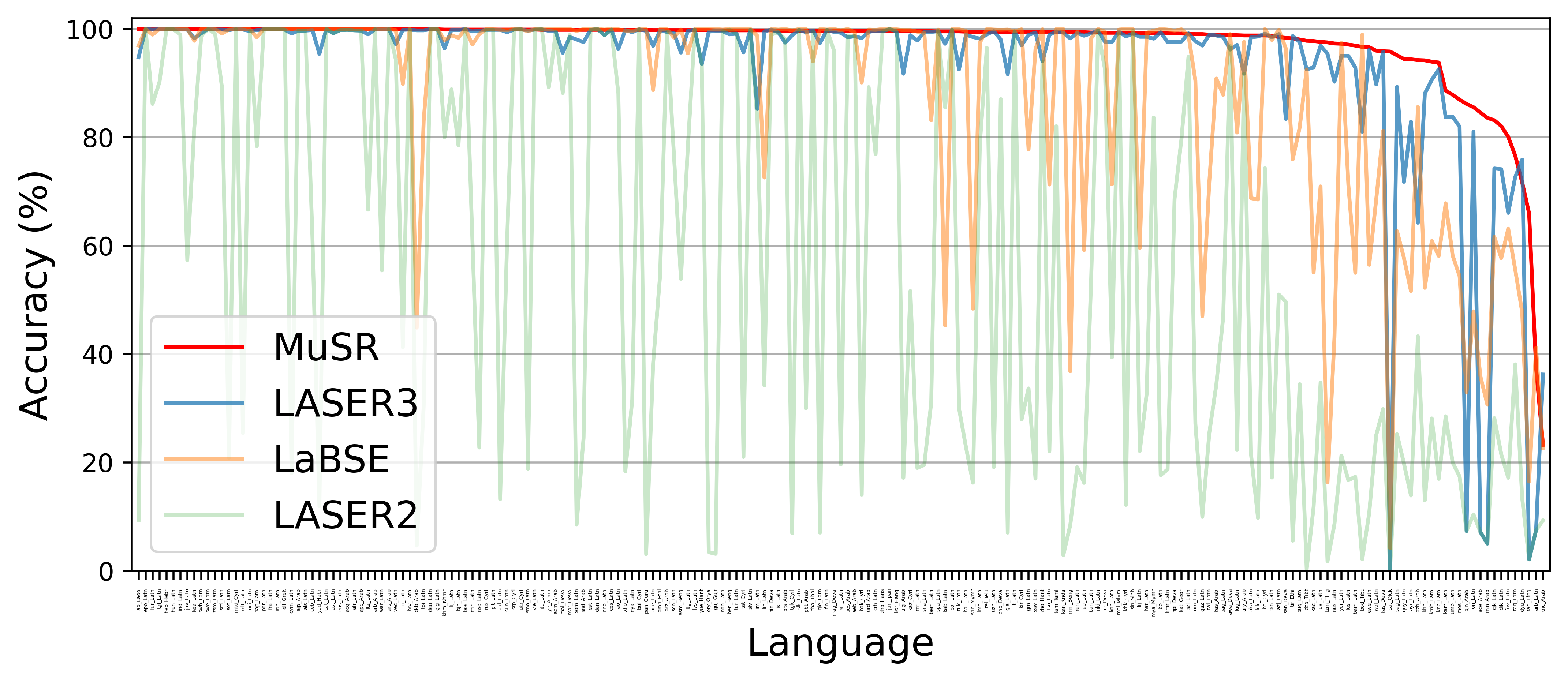}
\caption{The distribution of the averaged bidirectional accuracy with English of the multilingual similarity search on the Flores-200 benchmark.}
\label{fig:flores200_mss}
\end{figure}

As discussed in \citet{heffernan-etal-2022-bitext}, Tatoeba is less reliable for evaluating multilingual sentence embeddings since it mainly contains very short sentences which can introduce a strong bias towards a particular model or training corpus. We here illustrate the distribution of the averaged bidirectional accuracy of the strong baselines and MuSR on the Flores-200 benchmark in Figure \ref{fig:flores200_mss}. We can see that our approach performs strongly across a wide range of languages, with over $150$ languages achieving a similarity search accuracy exceeding $99\%$. LASER2 shows high variance across languages, and it could be resolved to some extent by incorporating language-specific models in LASER3.

% \begin{table*}[h]\small
% \centering
% \begin{tabular}{l | c c c c | c | c c c c c} 
% \multicolumn{1}{c|}{Method} & $L_e$ & $L_d$ & $D$ & $H$ & $P_e$ & \multicolumn{1}{c}{Tatoeba} & \multicolumn{2}{c}{Flores-100} & \multicolumn{2}{c}{Flores-200} \\
% & & & & & & $\leftrightarrow$ \texttt{en} & $\leftrightarrow$ \texttt{en} &  $\leftrightarrow$ \texttt{zh} & $\leftrightarrow$ \texttt{en} &  $\leftrightarrow$ \texttt{zh} \\
% \hline
% \hline
% NMT Pretrain & 256 \\
% \ \ + CrossConST & 256 \\
% \hline
% Multilingual NMT Pretrain & 12 & 3 & 512 & 8 & 252M & 78.89 & 98.55 & 98.09 & 95.30 & 94.38 \\
% \ \ + CrossConST Finetune & 12 & 3 & 512 & 8 & 252M & 82.69 & 98.94 & 98.24 & 96.25 & 94.76 \\
% \hline
% Multilingual NMT Pretrain & 12 & 3 & 768 & 12 & 434M & 80.76 & 98.86 & 98.33 & 96.36 & 95.33 \\
% \ \ + CrossConST Finetune & 12 & 3 & 768 & 12 & 434M & 83.96 & 99.23 & 98.48 & 97.37 & 95.95 \\
% \hline
% Multilingual NMT Pretrain & 12 & 3 & 1024 & 16 & 655M & 81.16 & 98.78 & 98.20 & 96.21 & 95.06 \\
% \ \ + CrossConST Finetune & 12 & 3 & 1024 & 16 & 655M & 84.25 & 99.07 & 98.34 & 97.29 & 96.02 \\
% \end{tabular}
% \caption{The averaged bidirectional similarity search accuracy according to different training stages and model architectures. $L_e$, $L_d$, $D$, $H$, and $P_e$ denotes the number of encoder layers, number of decoder layers, sentence embedding dimension, number of attention heads, and number of encoder parameters.}
% \label{tab:analysis}
% \end{table*}

\subsection{Bitext Mining}

Given two comparable corpora in different languages, we identify the sentence pairs that are translations of each other by leveraging the score \cite{artetxe-schwenk-2019-margin} defined as follows:
\begin{equation}\label{score}
\frac{\text{cos}(\mathbf{x}, \mathbf{y})}{\sum_{\mathbf{z} \in \text{NN}_k(\mathbf{x})}\frac{\text{cos}(\mathbf{x}, \mathbf{z})}{2k} + \sum_{\mathbf{z} \in \text{NN}_k(\mathbf{y})}\frac{\text{cos}(\mathbf{y}, \mathbf{z})}{2k}},
\end{equation}
where $\mathbf{x}$ and $\mathbf{y}$ are the source and target sentence embeddings respectively, and $\text{NN}_k(\mathbf{x})$ denotes the $k$ nearest neighbors of $\mathbf{x}$ in the other languages. We score each sentence pair by calculating \eqref{score}, and the parallel sentences are extracted and filtered by setting a fixed threshold over this score.

We conduct experiments on the BUCC dataset \cite{zweigenbaum2018overview} containing comparable corpora between English and four other languages: German (\texttt{de}), French (\texttt{fr}), Russian (\texttt{ru}), and Chinese (\texttt{zh}), using exact same hyperparameters as \citet{artetxe-schwenk-2019-margin}.\footnote{\url{https://github.com/facebookresearch/LASER/tree/main/tasks/bucc}} Given the monolingual corpora and the gold translation pairs, we extract the translation pairs from the monolingual data and evaluate against the ground truth. Following \citet{feng-etal-2022-language}, we evaluate the performance by F1 score on the training dataset since the ground truth for the test dataset is not released.

\begin{table}[h]\small
\centering
\begin{tabular}{c | c c c c | c} 
% \multicolumn{1}{c|}{Model} & \multicolumn{5}{c}{BUCC} \\
Model & \texttt{de} & \texttt{fr} & \texttt{ru} & \texttt{zh} & avg. \\
\hline
\hline
LASER2 & \bf 95.36 & 92.15 & 91.95 & 91.07 & 92.63 \\
LaBSE & \bf 95.86 & \bf 92.52 & \bf 92.46 & \bf 92.99 & \bf 93.46 \\
LASER3 & \bf 95.36 & 92.15 & 91.95 & 91.07 & 92.63 \\
\hline
MuSR & 94.91 & \bf 92.66 & \bf 92.25 & \bf 92.94 & \bf 93.19 \\
\end{tabular}
\caption{Our approach achieves the superior or comparable performance over the existing models on the BUCC benchmark. Note that LASER2 and LASER3 share the same model for the tested languages. We mark the best two scores in bold.}
\label{bucc_result}
\end{table}

We report the F1 scores of the strong baselines and our approach in Table \ref{bucc_result}. We can see that MuSR achieves strong performance on the bitext mining task. It is worth noting that all models perform similarly on the BUCC benchmark since the tested languages are all high resource languages. Our model however covers much more languages within a single model than LASER2 and LaBSE.

\subsection{Analysis}

\begin{table}[h]\small
\centering
\begin{tabular}{l | c c | c c c} 
\multicolumn{1}{c|}{Method} & $D$ & $H$ & \multicolumn{1}{c}{Tatoeba} & \multicolumn{2}{c}{Flores-200} \\
& & & $\leftrightarrow$ \texttt{en} & $\leftrightarrow$ \texttt{en} &  $\leftrightarrow$ \texttt{zh} \\
\hline
\hline
Phase 1 & 512 & 8 & 78.89 & 95.30 & 94.38 \\
Phase 2 & 512 & 8 & 82.69 & 96.25 & 94.76 \\
\hline
Phase 1 & 768 & 12 & 80.76 & 96.36 & 95.33 \\
Phase 2 & 768 & 12 & 83.96 & 97.37 & 95.95 \\
\hline
Phase 1 & 1024 & 16 & 81.16 & 96.21 & 95.06 \\
Phase 2 & 1024 & 16 & 84.25 & 97.29 & 96.02 \\
\end{tabular}
\caption{The averaged bidirectional similarity search accuracy according to different training stages and model architectures. $D$ and $H$ denote the sentence embedding dimension and the number of attention heads. Phase 1 denotes the multilingual NMT pretraining, and Phase 2 denotes the CrossConST finetuning.}
\label{tab:analysis}
\end{table}

We here investigate the impact of the cross-lingual consistency regularization and the model architectures on learning MuSR. We keep the training configurations the same except for the sentence embedding dimension and the number of attention heads. The experimental results on multilingual similarity search are summarized in Table \ref{tab:analysis}. By checking model performance under different combinations of training stage and architecture, we have the following observations: 1) The sentence representation model with multilingual NMT pretraining could achieve decent performance for non-English alignment, and CrossConST finetuning further boosts the model performance especially for English alignment. 2) The model performance consistently improves with the increasing of the sentence embedding dimension and the number of attention heads, while the models with $768$ and $1024$ embedding dimensions perform similarly, which is in line with \citet{feng-etal-2022-language}. Consider the computationally-heavy inference introduced by 655M parameters of the 1024-dim model, we choose 768 as the sentence embedding dimension.

\section{Conclusion}

In this paper, we propose MuSR: a one-for-all multilingual sentence representation model supporting $223$ languages. Experimental results show that MuSR could yield strong performance on various bitext retrieval and mining tasks compare with the SOTA models LaBSE and LASER3, while also providing increased language coverage in a single model. Extensive analysis shows that CrossConST and the sentence embedding dimension play the key roles in learning multilingual sentence representations. As for future work, we could explore the development of lightweight models by distilling knowledge from MuSR for multilingual sentence alignment, which would potentially lower the computational requirements and make the model more accessible for a variety of applications.

% \section*{Limitations}

% \section*{Ethics Statement}

% \section*{Acknowledgements}

\bibliography{anthology,custom}

\begin{thebibliography}{28}
\expandafter\ifx\csname natexlab\endcsname\relax\def\natexlab#1{#1}\fi

\bibitem[{Arivazhagan et~al.(2019)Arivazhagan, Bapna, Firat, Lepikhin, Johnson,
  Krikun, Chen, Cao, Foster, Cherry et~al.}]{arivazhagan2019massively}
Naveen Arivazhagan, Ankur Bapna, Orhan Firat, Dmitry Lepikhin, Melvin Johnson,
  Maxim Krikun, Mia~Xu Chen, Yuan Cao, George Foster, Colin Cherry, et~al.
  2019.
\newblock Massively multilingual neural machine translation in the wild:
  Findings and challenges.
\newblock \emph{arXiv preprint arXiv:1907.05019}.

\bibitem[{Artetxe and
  Schwenk(2019{\natexlab{a}})}]{artetxe-schwenk-2019-margin}
Mikel Artetxe and Holger Schwenk. 2019{\natexlab{a}}.
\newblock \href {https://doi.org/10.18653/v1/P19-1309} {Margin-based parallel
  corpus mining with multilingual sentence embeddings}.
\newblock In \emph{Proceedings of the 57th Annual Meeting of the Association
  for Computational Linguistics}, pages 3197--3203, Florence, Italy.
  Association for Computational Linguistics.

\bibitem[{Artetxe and
  Schwenk(2019{\natexlab{b}})}]{artetxe-schwenk-2019-massively}
Mikel Artetxe and Holger Schwenk. 2019{\natexlab{b}}.
\newblock \href {https://doi.org/10.1162/tacl_a_00288} {Massively multilingual
  sentence embeddings for zero-shot cross-lingual transfer and beyond}.
\newblock \emph{Transactions of the Association for Computational Linguistics},
  7:597--610.

\bibitem[{Ba{\~n}{\'o}n et~al.(2020)Ba{\~n}{\'o}n, Chen, Haddow, Heafield,
  Hoang, Espl{\`a}-Gomis, Forcada, Kamran, Kirefu, Koehn, Ortiz~Rojas,
  Pla~Sempere, Ram{\'\i}rez-S{\'a}nchez, Sarr{\'\i}as, Strelec, Thompson,
  Waites, Wiggins, and Zaragoza}]{banon-etal-2020-paracrawl}
Marta Ba{\~n}{\'o}n, Pinzhen Chen, Barry Haddow, Kenneth Heafield, Hieu Hoang,
  Miquel Espl{\`a}-Gomis, Mikel~L. Forcada, Amir Kamran, Faheem Kirefu, Philipp
  Koehn, Sergio Ortiz~Rojas, Leopoldo Pla~Sempere, Gema
  Ram{\'\i}rez-S{\'a}nchez, Elsa Sarr{\'\i}as, Marek Strelec, Brian Thompson,
  William Waites, Dion Wiggins, and Jaume Zaragoza. 2020.
\newblock \href {https://doi.org/10.18653/v1/2020.acl-main.417} {{P}ara{C}rawl:
  Web-scale acquisition of parallel corpora}.
\newblock In \emph{Proceedings of the 58th Annual Meeting of the Association
  for Computational Linguistics}, pages 4555--4567, Online. Association for
  Computational Linguistics.

\bibitem[{Bapna and Firat(2019)}]{bapna-firat-2019-simple}
Ankur Bapna and Orhan Firat. 2019.
\newblock \href {https://doi.org/10.18653/v1/D19-1165} {Simple, scalable
  adaptation for neural machine translation}.
\newblock In \emph{Proceedings of the 2019 Conference on Empirical Methods in
  Natural Language Processing and the 9th International Joint Conference on
  Natural Language Processing (EMNLP-IJCNLP)}, pages 1538--1548, Hong Kong,
  China. Association for Computational Linguistics.

\bibitem[{Christodouloupoulos and
  Steedman(2015)}]{christodouloupoulos2015massively}
Christos Christodouloupoulos and Mark Steedman. 2015.
\newblock A massively parallel corpus: the bible in 100 languages.
\newblock \emph{Language resources and evaluation}, 49:375--395.

\bibitem[{Conneau and Lample(2019)}]{NEURIPS2019_c04c19c2}
Alexis Conneau and Guillaume Lample. 2019.
\newblock \href
  {https://proceedings.neurips.cc/paper_files/paper/2019/file/c04c19c2c2474dbf5f7ac4372c5b9af1-Paper.pdf}
  {Cross-lingual language model pretraining}.
\newblock In \emph{Advances in Neural Information Processing Systems},
  volume~32. Curran Associates, Inc.

\bibitem[{Costa-juss{\`a} et~al.(2022)Costa-juss{\`a}, Cross, {\c{C}}elebi,
  Elbayad, Heafield, Heffernan, Kalbassi, Lam, Licht, Maillard
  et~al.}]{costa2022no}
Marta~R Costa-juss{\`a}, James Cross, Onur {\c{C}}elebi, Maha Elbayad, Kenneth
  Heafield, Kevin Heffernan, Elahe Kalbassi, Janice Lam, Daniel Licht, Jean
  Maillard, et~al. 2022.
\newblock No language left behind: Scaling human-centered machine translation.
\newblock \emph{arXiv preprint arXiv:2207.04672}.

\bibitem[{El-Kishky et~al.(2020)El-Kishky, Chaudhary, Guzm{\'a}n, and
  Koehn}]{el-kishky-etal-2020-ccaligned}
Ahmed El-Kishky, Vishrav Chaudhary, Francisco Guzm{\'a}n, and Philipp Koehn.
  2020.
\newblock \href {https://doi.org/10.18653/v1/2020.emnlp-main.480} {{CCA}ligned:
  A massive collection of cross-lingual web-document pairs}.
\newblock In \emph{Proceedings of the 2020 Conference on Empirical Methods in
  Natural Language Processing (EMNLP)}, pages 5960--5969, Online. Association
  for Computational Linguistics.

\bibitem[{Feng et~al.(2022)Feng, Yang, Cer, Arivazhagan, and
  Wang}]{feng-etal-2022-language}
Fangxiaoyu Feng, Yinfei Yang, Daniel Cer, Naveen Arivazhagan, and Wei Wang.
  2022.
\newblock \href {https://doi.org/10.18653/v1/2022.acl-long.62}
  {Language-agnostic {BERT} sentence embedding}.
\newblock In \emph{Proceedings of the 60th Annual Meeting of the Association
  for Computational Linguistics (Volume 1: Long Papers)}, pages 878--891,
  Dublin, Ireland. Association for Computational Linguistics.

\bibitem[{Gao et~al.(2023)Gao, Zhang, He, Wu, and Wang}]{gao2023crossconst}
Pengzhi Gao, Liwen Zhang, Zhongjun He, Hua Wu, and Haifeng Wang. 2023.
\newblock Improving zero-shot multilingual neural machine translation by
  leveraging cross-lingual consistency regularization.
\newblock \emph{arXiv preprint arXiv:2305.07310}.

\bibitem[{Guo et~al.(2018)Guo, Shen, Yang, Ge, Cer, Hernandez~Abrego, Stevens,
  Constant, Sung, Strope, and Kurzweil}]{guo-etal-2018-effective}
Mandy Guo, Qinlan Shen, Yinfei Yang, Heming Ge, Daniel Cer, Gustavo
  Hernandez~Abrego, Keith Stevens, Noah Constant, Yun-Hsuan Sung, Brian Strope,
  and Ray Kurzweil. 2018.
\newblock \href {https://doi.org/10.18653/v1/W18-6317} {Effective parallel
  corpus mining using bilingual sentence embeddings}.
\newblock In \emph{Proceedings of the Third Conference on Machine Translation:
  Research Papers}, pages 165--176, Brussels, Belgium. Association for
  Computational Linguistics.

\bibitem[{Heffernan et~al.(2022)Heffernan, {\c{C}}elebi, and
  Schwenk}]{heffernan-etal-2022-bitext}
Kevin Heffernan, Onur {\c{C}}elebi, and Holger Schwenk. 2022.
\newblock \href {https://aclanthology.org/2022.findings-emnlp.154} {Bitext
  mining using distilled sentence representations for low-resource languages}.
\newblock In \emph{Findings of the Association for Computational Linguistics:
  EMNLP 2022}, pages 2101--2112, Abu Dhabi, United Arab Emirates. Association
  for Computational Linguistics.

\bibitem[{Joulin et~al.(2016)Joulin, Grave, Bojanowski, Douze, J{\'e}gou, and
  Mikolov}]{joulin2016fasttext}
Armand Joulin, Edouard Grave, Piotr Bojanowski, Matthijs Douze, H{\'e}rve
  J{\'e}gou, and Tomas Mikolov. 2016.
\newblock Fasttext.zip: Compressing text classification models.
\newblock \emph{arXiv preprint arXiv:1612.03651}.

\bibitem[{Joulin et~al.(2017)Joulin, Grave, Bojanowski, and
  Mikolov}]{joulin-etal-2017-bag}
Armand Joulin, Edouard Grave, Piotr Bojanowski, and Tomas Mikolov. 2017.
\newblock \href {https://aclanthology.org/E17-2068} {Bag of tricks for
  efficient text classification}.
\newblock In \emph{Proceedings of the 15th Conference of the {E}uropean Chapter
  of the Association for Computational Linguistics: Volume 2, Short Papers},
  pages 427--431, Valencia, Spain. Association for Computational Linguistics.

\bibitem[{Junczys-Dowmunt(2018)}]{junczys-dowmunt-2018-dual}
Marcin Junczys-Dowmunt. 2018.
\newblock \href {https://doi.org/10.18653/v1/W18-6478} {Dual conditional
  cross-entropy filtering of noisy parallel corpora}.
\newblock In \emph{Proceedings of the Third Conference on Machine Translation:
  Shared Task Papers}, pages 888--895, Belgium, Brussels. Association for
  Computational Linguistics.

\bibitem[{Kudo and Richardson(2018)}]{kudo-richardson-2018-sentencepiece}
Taku Kudo and John Richardson. 2018.
\newblock \href {https://doi.org/10.18653/v1/D18-2012} {{S}entence{P}iece: A
  simple and language independent subword tokenizer and detokenizer for neural
  text processing}.
\newblock In \emph{Proceedings of the 2018 Conference on Empirical Methods in
  Natural Language Processing: System Demonstrations}, pages 66--71, Brussels,
  Belgium. Association for Computational Linguistics.

\bibitem[{Mao and Nakagawa(2023)}]{mao-nakagawa-2023-lealla}
Zhuoyuan Mao and Tetsuji Nakagawa. 2023.
\newblock \href {https://aclanthology.org/2023.eacl-main.138} {{LEALLA}:
  Learning lightweight language-agnostic sentence embeddings with knowledge
  distillation}.
\newblock In \emph{Proceedings of the 17th Conference of the European Chapter
  of the Association for Computational Linguistics}, pages 1886--1894,
  Dubrovnik, Croatia. Association for Computational Linguistics.

\bibitem[{Reimers and Gurevych(2020)}]{reimers-gurevych-2020-making}
Nils Reimers and Iryna Gurevych. 2020.
\newblock \href {https://doi.org/10.18653/v1/2020.emnlp-main.365} {Making
  monolingual sentence embeddings multilingual using knowledge distillation}.
\newblock In \emph{Proceedings of the 2020 Conference on Empirical Methods in
  Natural Language Processing (EMNLP)}, pages 4512--4525, Online. Association
  for Computational Linguistics.

\bibitem[{Schwenk et~al.(2021{\natexlab{a}})Schwenk, Chaudhary, Sun, Gong, and
  Guzm{\'a}n}]{schwenk-etal-2021-wikimatrix}
Holger Schwenk, Vishrav Chaudhary, Shuo Sun, Hongyu Gong, and Francisco
  Guzm{\'a}n. 2021{\natexlab{a}}.
\newblock \href {https://doi.org/10.18653/v1/2021.eacl-main.115}
  {{W}iki{M}atrix: Mining 135{M} parallel sentences in 1620 language pairs from
  {W}ikipedia}.
\newblock In \emph{Proceedings of the 16th Conference of the European Chapter
  of the Association for Computational Linguistics: Main Volume}, pages
  1351--1361, Online. Association for Computational Linguistics.

\bibitem[{Schwenk et~al.(2021{\natexlab{b}})Schwenk, Wenzek, Edunov, Grave,
  Joulin, and Fan}]{schwenk-etal-2021-ccmatrix}
Holger Schwenk, Guillaume Wenzek, Sergey Edunov, Edouard Grave, Armand Joulin,
  and Angela Fan. 2021{\natexlab{b}}.
\newblock \href {https://doi.org/10.18653/v1/2021.acl-long.507} {{CCM}atrix:
  Mining billions of high-quality parallel sentences on the web}.
\newblock In \emph{Proceedings of the 59th Annual Meeting of the Association
  for Computational Linguistics and the 11th International Joint Conference on
  Natural Language Processing (Volume 1: Long Papers)}, pages 6490--6500,
  Online. Association for Computational Linguistics.

\bibitem[{Sennrich et~al.(2016)Sennrich, Haddow, and
  Birch}]{sennrich-etal-2016-neural}
Rico Sennrich, Barry Haddow, and Alexandra Birch. 2016.
\newblock \href {https://doi.org/10.18653/v1/P16-1162} {Neural machine
  translation of rare words with subword units}.
\newblock In \emph{Proceedings of the 54th Annual Meeting of the Association
  for Computational Linguistics (Volume 1: Long Papers)}, pages 1715--1725,
  Berlin, Germany. Association for Computational Linguistics.

\bibitem[{Tiedemann(2016)}]{tiedemann-2016-finding}
J{\"o}rg Tiedemann. 2016.
\newblock \href {https://aclanthology.org/L16-1559} {Finding alternative
  translations in a large corpus of movie subtitle}.
\newblock In \emph{Proceedings of the Tenth International Conference on
  Language Resources and Evaluation ({LREC}'16)}, pages 3518--3522,
  Portoro{\v{z}}, Slovenia. European Language Resources Association (ELRA).

\bibitem[{Tiedemann(2012)}]{TIEDEMANN12.463}
Jörg Tiedemann. 2012.
\newblock Parallel data, tools and interfaces in opus.
\newblock In \emph{Proceedings of the Eight International Conference on
  Language Resources and Evaluation (LREC'12)}, Istanbul, Turkey. European
  Language Resources Association (ELRA).

\bibitem[{Vaswani et~al.(2017)Vaswani, Shazeer, Parmar, Uszkoreit, Jones,
  Gomez, Kaiser, and Polosukhin}]{vaswani2017attention}
Ashish Vaswani, Noam Shazeer, Niki Parmar, Jakob Uszkoreit, Llion Jones,
  Aidan~N. Gomez, Lukasz Kaiser, and Illia Polosukhin. 2017.
\newblock \href
  {https://proceedings.neurips.cc/paper/2017/hash/3f5ee243547dee91fbd053c1c4a845aa-Abstract.html}
  {Attention is all you need}.
\newblock In \emph{Advances in Neural Information Processing Systems 30: Annual
  Conference on Neural Information Processing Systems 2017, December 4-9, 2017,
  Long Beach, CA, {USA}}, pages 5998--6008.

\bibitem[{Yang et~al.(2020)Yang, Cer, Ahmad, Guo, Law, Constant,
  Hernandez~Abrego, Yuan, Tar, Sung, Strope, and
  Kurzweil}]{yang-etal-2020-multilingual}
Yinfei Yang, Daniel Cer, Amin Ahmad, Mandy Guo, Jax Law, Noah Constant, Gustavo
  Hernandez~Abrego, Steve Yuan, Chris Tar, Yun-hsuan Sung, Brian Strope, and
  Ray Kurzweil. 2020.
\newblock \href {https://doi.org/10.18653/v1/2020.acl-demos.12} {Multilingual
  universal sentence encoder for semantic retrieval}.
\newblock In \emph{Proceedings of the 58th Annual Meeting of the Association
  for Computational Linguistics: System Demonstrations}, pages 87--94, Online.
  Association for Computational Linguistics.

\bibitem[{Yang et~al.(2019)Yang, Hernandez~Abrego, Yuan, Guo, Shen, Cer, Sung,
  Strope, and Kurzweil}]{ijcai2019p746}
Yinfei Yang, Gustavo Hernandez~Abrego, Steve Yuan, Mandy Guo, Qinlan Shen,
  Daniel Cer, Yun-hsuan Sung, Brian Strope, and Ray Kurzweil. 2019.
\newblock \href {https://doi.org/10.24963/ijcai.2019/746} {Improving
  multilingual sentence embedding using bi-directional dual encoder with
  additive margin softmax}.
\newblock In \emph{Proceedings of the Twenty-Eighth International Joint
  Conference on Artificial Intelligence, {IJCAI-19}}, pages 5370--5378.
  International Joint Conferences on Artificial Intelligence Organization.

\bibitem[{Zweigenbaum et~al.(2018)Zweigenbaum, Sharoff, and
  Rapp}]{zweigenbaum2018overview}
Pierre Zweigenbaum, Serge Sharoff, and Reinhard Rapp. 2018.
\newblock Overview of the third bucc shared task: Spotting parallel sentences
  in comparable corpora.
\newblock In \emph{Proceedings of 11th workshop on building and using
  comparable corpora}, pages 39--42.

\end{thebibliography}
\bibliographystyle{acl_natbib}

\appendix

\section*{Appendix}

\begin{table*}[h]\small
\centering
\begin{tabular}{c | c | c | c} 
Language & Language & Language & Language \\
\hline
\hline
Acehnese (Arabic script) & Georgian & Mossi & Tsonga \\
Acehnese (Latin script) & German & Najdi Arabic & Tswana \\
Afrikaans & Greek & Nepali & Tumbuka \\
Akan & Guarani & Nigerian Fulfulde & Tunisian Arabic \\
Algerian Arabic & Gujarati & North Azerbaijani & Turkish \\
Amharic & Haitian Creole & North Levantine Arabic & Turkmen \\
Armenian & Halh Mongolian & Northern Kurdish & Twi \\
Assamese & Hausa & Northern Sotho & Ukrainian \\
Asturian & Hebrew & Northern Uzbek & Umbundu \\
Awadhi & Hindi & Norwegian Bokmål & Upper Sorbian \\
Ayacucho Quechua & Hungarian & Norwegian Nynorsk & Urdu \\
Balinese & Icelandic & Nuer & Uyghur \\
Bambara & Ido & Nyanja & Venetian \\
Banjar (Arabic script) & Igbo & Occitan & Vietnamese \\
Banjar (Latin script) & Ilocano / Iloko & Odia & Walloon \\
Bashkir & Indonesian & Pangasinan & Waray \\
Basque & Interlingua & Papiamento & Welsh \\
Belarusian & Interlingue & Plateau Malagasy & West Central Oromo \\
Bemba & Irish & Polish & Western Frisian \\
Bengali & Italian & Portuguese & Western Persian \\
Berber languages & Japanese & Romanian & Wolof \\
Bhojpuri & Javanese & Rundi & Xhosa \\
Bosnian & Jingpho & Russian & Yoruba \\
Breton & Kabiyè & Samoan & Yue Chinese \\
Buginese & Kabuverdianu & Sango & Zulu \\
Bulgarian & Kabyle & Sanskrit \\
Burmese & Kamba & Santali \\
Catalan & Kannada & Sardinian \\
Cebuano & Kashmiri (Arabic script) & Scottish Gaelic \\
Central Atlas Tamazight & Kashmiri (Devanagari script) & Serbian \\
Central Aymara & Kashubian & Serbo-Croatian \\
Central Kanuri (Arabic script) & Kazakh & Shan \\
Central Kanuri (Latin script) & Khmer & Shanghainese \\
Central Kurdish & Kikongo & Shona \\
Chamorro & Kikuyu & Sicilian \\
Chhattisgarhi & Kimbundu & Silesian \\
Chinese (Simplified) & Kinyarwanda & Sindhi \\
Chinese (Traditional) & Korean & Sinhala \\
Chokwe & Kyrgyz & Slovak \\
Chuvash & Lao & Slovenian \\
Cornish & Latgalian & Somali \\
Crimean Tatar & Latin & South Azerbaijani \\
Croatian & Ligurian & South Levantine Arabic \\
Czech & Limburgish & Southern Pashto \\
Danish & Lingala & Southern Sotho \\
Dari & Lingua Franca Nova & Southwestern Dinka \\
Divehi & Lithuanian & Spanish \\
Dutch & Lojban & Standard Latvian \\
Dyula & Lombard & Standard Malay \\
Dzongkha & Low German & Standard Tibetan \\
Eastern Panjabi & Luba-Kasai & Sundanese \\
Eastern Yiddish & Luo & Swahili \\
Egyptian Arabic & Luxembourgish & Swati \\
English & Macedonian & Swedish \\
Esperanto & Magahi & Tagalog \\
Estonian & Maithili & Tajik \\
Ewe & Malayalam & Tamasheq (Latin script) \\
Faroese & Maltese & Tamasheq (Tifinagh script) \\
Fijian & Maori & Tamil \\
Filipino & Marathi & Tatar \\
Finnish & Meitei (Bengali script) & Ta’izzi-Adeni Arabic \\
Fon & Mesopotamian Arabic & Telugu \\
French & Minangkabau (Latin script) & Thai \\
Friulian & Mizo & Tigrinya \\
Galician & Modern Standard Arabic & Tok Pisin \\
Ganda & Moroccan Arabic & Tosk Albanian \\
\hline
\end{tabular}
\caption{The supported languages of MuSR.}
\label{tab:langs}
\end{table*}

\begin{figure*}[h]
\centering
\includegraphics[scale=0.99]{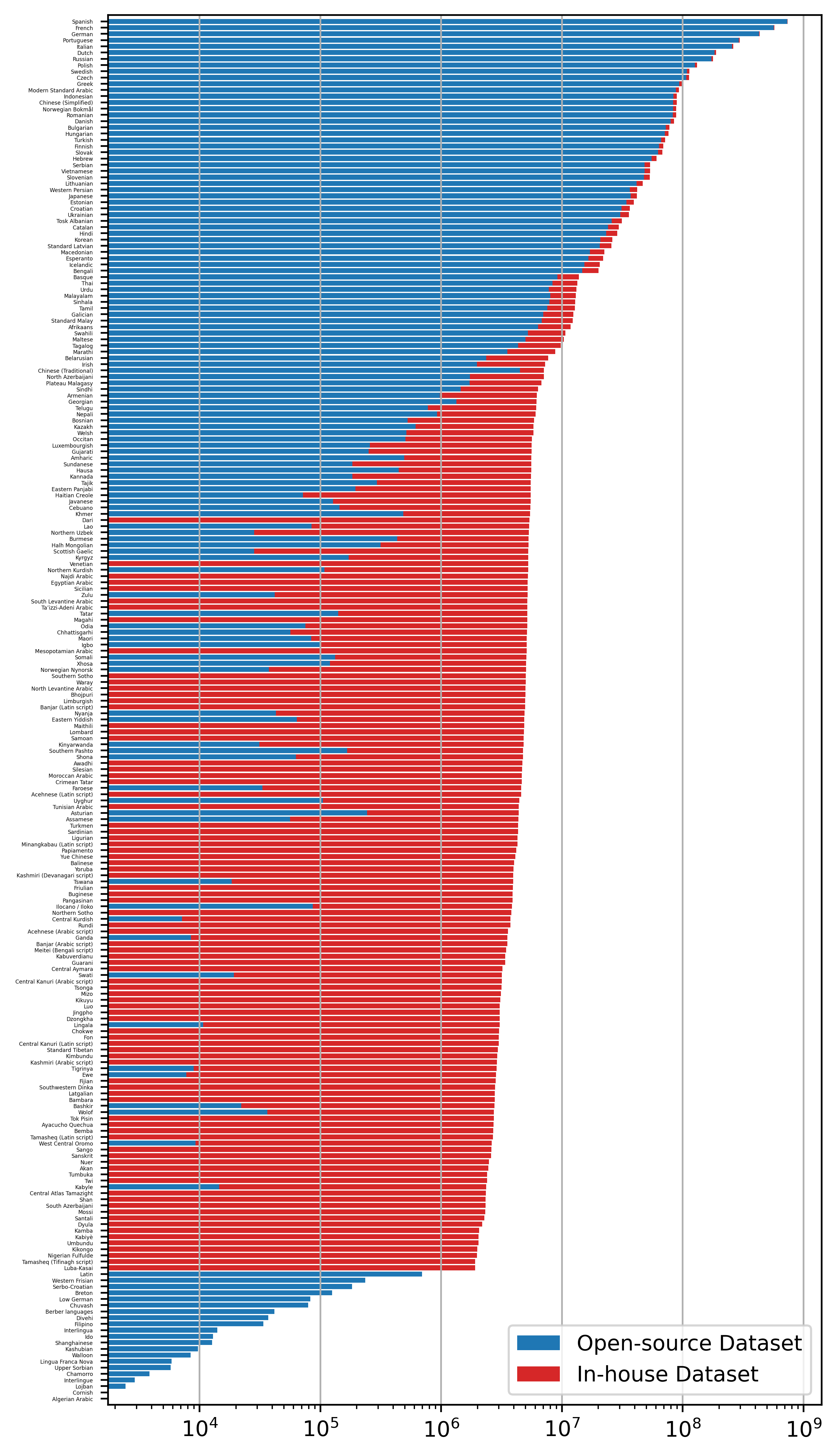}
\caption{The distribution of the open-source and in-house cleaned datasets for each language in our training dataset.}
\label{fig:distribution_full}
\end{figure*}

\begin{table*}[h]\small
\centering
\begin{tabular}{c | c c c c | c | c c c c} 
Language & LASER2 & LASER3 & LaBSE & MuSR & Language & LASER2 & LASER3 & LaBSE & MuSR \\
\hline
\hline
afr & 93.2 & - & \bf 97.4 & 95.85 & kaz & 55.83 & 80.61 & \bf 90.52 & 87.48 \\
amh & 80.06 & 86.31 & \bf 94.05 & 88.39 & khm & 77.49 & 53.32 & \bf 83.17 & 77.35 \\
ang & 37.31 & - & \bf 64.55 & 57.84 & kor & 91.35 & - & \bf 93.5 & 89.9 \\
ara & \bf 92.25 & - & 90.85 & 90.45 & kur & 23.41 & - & \bf 87.2 & 78.54 \\
arq & 33.04 & - & 46.16 & \bf 65.59 & kzj & 8.65 & - & \bf 14.25 & 13.95 \\
arz & 70.02 & - & 78.41 & \bf 82.39 & lat & 68.9 & - & \bf 81.9 & 70.5 \\
ast & 80.71 & - & \bf 90.55 & 90.16 & lfn & 67.85 & - & 71.25 & \bf 84.9 \\
awa & 39.39 & 80.74 & 73.16 & \bf 85.93 & lit & 96.95 & - & \bf 97.3 & 95.8 \\
aze & 81.65 & 91.5 & \bf 96.1 & 92.95 & lvs & 96.6 & - & \bf 96.8 & 94.7 \\
bel & 83.4 & 94.05 & \bf 96.15 & 95.05 & mal & 98.4 & 97.82 & \bf 98.91 & 97.67 \\
ben & 91.3 & 90.1 & \bf 91.35 & 89.4 & mar & \bf 94.75 & 91.1 & 94.7 & 94.5 \\
ber & \bf 81.75 & - & 10.5 & 74.7 & max & 45.42 & - & \bf 71.13 & 66.02 \\
bos & \bf 96.89 & - & 96.33 & 96.75 & mhr & 10 & - & \bf 19.5 & 12.3 \\
bre & \bf 36.6 & - & 17.35 & 21.65 & mkd & \bf 95.1 & - & 94.85 & 94.65 \\
bul & 95.15 & - & \bf 95.7 & 95.05 & mon & 7.27 & 87.73 & \bf 96.48 & 88.52 \\
cat & 96.55 & - & \bf 96.6 & 96.25 & nds & 80.2 & - & 81.35 & \bf 88.75 \\
cbk & 79.75 & - & \bf 82.4 & 77.2 & nld & 96.35 & - & \bf 97.25 & 96.45 \\
ceb & 15.92 & \bf 80 & 71 & 62.17 & nno & 77.25 & - & 95.85 & \bf 96 \\
ces & 96.85 & - & \bf 97.5 & 96.25 & nob & 95.6 & - & \bf 98.9 & 98.5 \\
cha & 26.64 & - & 39.05 & \bf 44.53 & nov & 67.51 & - & 78.21 & \bf 85.02 \\
cmn & 84.3 & - & \bf 96.2 & 94.85 & oci & 63.35 & - & 69.75 & \bf 76.85 \\
cor & 7.2 & - & 12.75 & \bf 24.95 & orv & 30.24 & - & \bf 47.07 & 44.01 \\
csb & 38.34 & - & 56.13 & \bf 66.21 & pam & 5.5 & - & \bf 13.55 & 13.2 \\
cym & 9.74 & 89.04 & \bf 93.65 & 87.22 & pes & 92.9 & 93.4 & \bf 96.05 & 94.45 \\
dan & 95.9 & - & \bf 96.45 & 96.25 & pms & 45.14 & - & 66.95 & \bf 86.67 \\
deu & 99.3 & - & \bf 99.35 & 98.95 & pol & \bf 98 & - & 97.85 & 97.85 \\
dsb & 51.25 & - & \bf 69.31 & 69 & por & \bf 95.75 & - & 95.55 & 95.4 \\
dtp & 11.5 & - & 13.35 & \bf 21.8 & ron & 97.25 & - & \bf 97.85 & 97.45 \\
ell & \bf 96.85 & - & 96.6 & 96.55 & rus & 94.35 & - & \bf 95.3 & 95 \\
epo & 97.45 & - & \bf 98.35 & 97.65 & slk & 96.6 & - & \bf 97.3 & 96.55 \\
est & 97 & - & \bf 97.7 & 96.45 & slv & \bf 96.78 & - & 96.72 & 95.63 \\
eus & 93.85 & - & \bf 95.75 & 94 & spa & 97.9 & - & \bf 98.45 & 97.75 \\
fao & 64.12 & 73.66 & 90.46 & \bf 93.32 & sqi & \bf 97.85 & \bf 97.85 & 97.65 & 97.05 \\
fin & \bf 97.3 & - & 97.05 & 95.85 & srp & 95.05 & - & \bf 96.2 & 95.9 \\
fra & 95.5 & - & \bf 96.05 & 95.6 & swe & 95.85 & - & \bf 96.55 & 96.45 \\
fry & 51.45 & - & \bf 90.17 & 71.97 & swg & 45.09 & - & \bf 65.18 & \bf 65.18 \\
gla & 3.32 & 70.27 & \bf 88.9 & 82.51 & swh & 57.69 & 81.41 & \bf 88.46 & 80.13 \\
gle & 9.15 & 78.55 & \bf 95 & 88.75 & tam & 85.99 & 58.79 & \bf 90.72 & 85.18 \\
glg & 96.75 & - & \bf 97.25 & 95.5 & tat & 30.7 & 64.7 & \bf 87.9 & 86.5 \\
gsw & 36.32 & - & 52.56 & \bf 66.67 & tel & 97.01 & 80.56 & \bf 98.29 & 92.31 \\
heb & 91.75 & - & \bf 92.95 & 91.85 & tgl & 68.85 & 95 & \bf 97.45 & 91.6 \\
hin & 96.1 & 95.55 & \bf 97.75 & 97.05 & tha & 96.99 & 96.53 & \bf 97.08 & 95.71 \\
hrv & 97.45 & - & \bf 97.8 & 97.5 & tuk & 22.17 & 58.37 & 80.05 & \bf 86.45 \\
hsb & 54.04 & - & 71.12 & \bf 80.43 & tur & 98.15 & 97.2 & \bf 98.35 & 97.85 \\
hun & 96.1 & - & \bf 97.2 & 96.15 & tzl & 41.35 & - & \bf 62.98 & 57.69 \\
hye & 90.03 & 90.63 & \bf 95.01 & 92.18 & uig & 51.45 & 76.3 & \bf 93.7 & 89.3 \\
ido & 84.1 & - & 90.8 & \bf 94.5 & ukr & 95.05 & - & \bf 95.25 & 95.1 \\
ile & 88.85 & - & 87.05 & \bf 95.85 & urd & 82.6 & 89.85 & \bf 95.35 & 92.55 \\
ina & 95.5 & - & 95.85 & \bf 96.75 & uzb & 26.4 & 78.39 & \bf 86.8 & 74.65 \\
ind & 94.8 & 94.75 & \bf 95.3 & 94.75 & vie & 97.15 & - & \bf 97.85 & 96.55 \\
isl & 95.8 & - & 96.15 & \bf 96.25 & war & 13.35 & \bf 75.35 & 65.4 & 70.1 \\
ita & \bf 95.55 & - & 94.65 & 95.25 & wuu & 79.4 & - & \bf 90.3 & 89.45 \\
jav & 18.78 & \bf 86.34 & 84.39 & 81.22 & xho & 5.63 & \bf 93.66 & 91.9 & 91.2 \\
jpn & 96 & - & \bf 96.45 & 94.35 & yid & 5.19 & \bf 94.16 & 90.98 & 89.86 \\
kab & 71.45 & \bf 89.65 & 6 & 72.55 & yue & 87.65 & - & \bf 92.1 & 86.35 \\
kat & 81.97 & 75 & \bf 95.91 & 93.43 & zsm & 96.25 & 96.1 & \bf 96.9 & 95.85 \\
\hline
\end{tabular}
\caption{The averaged bidirectional similarity search accuracy (\texttt{xx} $\leftrightarrow$ English) on the Tatoeba benchmark.}
\label{tab:tatoeba}
\end{table*}

\begin{table*}[h]\small
\centering
\begin{tabular}{c | c c c c | c | c c c c} 
Language & LASER2 & LASER3 & LaBSE & MuSR & Language & LASER2 & LASER3 & LaBSE & MuSR \\
\hline
\hline
ace\_Arab & 7.11 & - & 35.82 & \bf 83.84 & gaz\_Latn & 9.93 & 96.94 & 46.99 & \bf 99.01 \\
ace\_Latn & 38.24 & 96.89 & 88.74 & \bf 99.6 & gla\_Latn & 7.02 & 91.65 & \bf 99.9 & 99.65 \\
acm\_Arab & 99.51 & - & \bf 100 & 99.9 & gle\_Latn & 7.02 & 97.38 & \bf 100 & 99.7 \\
acq\_Arab & 99.85 & - & \bf 100 & \bf 100 & glg\_Latn & 99.95 & - & \bf 100 & 99.95 \\
aeb\_Arab & 98.67 & - & 99.41 & \bf 99.65 & grn\_Latn & 33.65 & 98.91 & 77.77 & \bf 99.31 \\
afr\_Latn & 99.75 & - & \bf 100 & 99.95 & guj\_Gujr & 3.11 & 99.65 & \bf 100 & 99.95 \\
ajp\_Arab & 99.7 & - & \bf 99.95 & \bf 99.95 & hat\_Latn & 32.71 & 98.57 & \bf 99.31 & 99.21 \\
aka\_Latn & 21.49 & 98.47 & 68.77 & \bf 99.06 & hau\_Latn & 22.78 & 98.96 & \bf 99.7 & 99.56 \\
als\_Latn & 99.7 & - & \bf 100 & \bf 100 & heb\_Hebr & 99.95 & - & \bf 100 & \bf 100 \\
amh\_Ethi & 54.5 & 99.75 & \bf 100 & 99.9 & hin\_Deva & 98.96 & 99.9 & \bf 100 & 99.85 \\
apc\_Arab & 99.7 & - & \bf 100 & 99.95 & hne\_Deva & 92.49 & 97.63 & \bf 99.51 & \bf 99.51 \\
arb\_Arab & 99.95 & - & \bf 100 & \bf 100 & hrv\_Latn & 99.9 & - & \bf 100 & 99.95 \\
arb\_Latn & 7.46 & - & \bf 41.16 & 35.52 & hun\_Latn & 99.95 & - & \bf 100 & \bf 100 \\
ars\_Arab & 99.95 & - & \bf 100 & \bf 100 & hye\_Armn & 89.23 & 99.65 & \bf 100 & 99.85 \\
ary\_Arab & 91.75 & - & 97.63 & \bf 98.81 & ibo\_Latn & 17.64 & 99.41 & \bf 100 & 99.65 \\
arz\_Arab & 99.46 & - & \bf 99.95 & 99.85 & ilo\_Latn & 41.25 & 99.85 & 89.87 & \bf 100 \\
asm\_Beng & 53.85 & 95.65 & \bf 99.9 & 99.75 & ind\_Latn & 98.96 & 99.9 & \bf 100 & \bf 100 \\
ast\_Latn & 99.21 & - & 99.95 & \bf 100 & isl\_Latn & 99.41 & - & \bf 99.9 & 99.75 \\
awa\_Deva & 96.89 & 96.2 & \bf 99.06 & 99.01 & ita\_Latn & 99.95 & - & \bf 100 & 99.9 \\
ayr\_Latn & 13.88 & 82.91 & 51.63 & \bf 94.47 & jav\_Latn & 57.31 & 99.9 & \bf 100 & 99.95 \\
azb\_Arab & 43.28 & 64.23 & 85.62 & \bf 93.82 & jpn\_Jpan & \bf 100 & - & \bf 100 & 99.7 \\
azj\_Latn & 50.99 & 99.06 & \bf 99.85 & 98.67 & kab\_Latn & 85.52 & 97.28 & 45.26 & \bf 99.26 \\
bak\_Cyrl & 13.98 & 98.32 & 90.12 & \bf 99.7 & kac\_Latn & 11.76 & 92.93 & 55.04 & \bf 98.22 \\
bam\_Latn & 17.34 & 92.89 & 54.99 & \bf 96.49 & kam\_Latn & 28.51 & 83.7 & 67.84 & \bf 86.91 \\
ban\_Latn & 53.46 & 99.21 & 98.27 & \bf 99.41 & kan\_Knda & 2.87 & 99.31 & \bf 100 & 99.7 \\
bel\_Cyrl & 74.31 & 99.16 & \bf 100 & 99.11 & kas\_Arab & 34.29 & 98.81 & 90.86 & \bf 99.01 \\
bem\_Latn & 31.03 & 99.46 & 83.15 & \bf 99.6 & kas\_Deva & 29.84 & \bf 95.8 & 81.23 & 95.06 \\
ben\_Beng & 99.9 & 99.01 & \bf 100 & 99.85 & kat\_Geor & 79.79 & 97.68 & \bf 99.95 & 99.36 \\
bho\_Deva & 87.06 & 98.07 & \bf 99.85 & 99.7 & kaz\_Cyrl & 51.63 & 98.86 & \bf 99.8 & 99.56 \\
bjn\_Arab & 7.31 & - & 32.91 & \bf 83.55 & kbp\_Latn & 12.99 & 88.09 & 52.22 & \bf 93.82 \\
bjn\_Latn & 78.51 & \bf 99.8 & 98.37 & \bf 99.8 & kea\_Latn & 81.67 & 98.27 & 97.83 & \bf 100 \\
bod\_Tibt & 2.12 & 81.03 & \bf 98.96 & 97.48 & khk\_Cyrl & 12.15 & 98.62 & \bf 100 & 99.51 \\
bos\_Latn & \bf 100 & - & \bf 100 & 99.9 & khm\_Khmr & 79.99 & 96.39 & 97.92 & \bf 99.95 \\
bug\_Latn & 34.44 & 97.58 & 81.82 & \bf 97.97 & kik\_Latn & 9.73 & \bf 98.62 & 68.53 & \bf 98.62 \\
bul\_Cyrl & 99.95 & - & \bf 100 & 99.75 & kin\_Latn & 19.61 & 99.31 & \bf 99.75 & \bf 99.75 \\
cat\_Latn & \bf 100 & - & \bf 100 & \bf 100 & kir\_Cyrl & 27.92 & 96.99 & \bf 99.95 & 99.11 \\
ceb\_Latn & 61.41 & 99.8 & \bf 100 & \bf 100 & kmb\_Latn & 28.11 & 90.61 & 60.87 & \bf 93.58 \\
ces\_Latn & 99.9 & - & \bf 100 & 99.9 & kmr\_Latn & 18.68 & 97.58 & \bf 99.9 & 99.51 \\
cjk\_Latn & 28.16 & 74.26 & 61.61 & \bf 82.31 & knc\_Arab & 9.29 & \bf 36.22 & 22.68 & 21.99 \\
ckb\_Arab & 4.64 & 99.75 & 44.86 & \bf 99.95 & knc\_Latn & 16.95 & 92.59 & 58.1 & \bf 93.13 \\
crh\_Latn & 76.88 & 99.7 & \bf 99.85 & 99.7 & kon\_Latn & 39.38 & 97.63 & 71.34 & \bf 99.26 \\
cym\_Latn & 18.03 & 99.16 & \bf 100 & \bf 100 & kor\_Hang & 99.56 & - & \bf 99.95 & 99.8 \\
dan\_Latn & \bf 100 & - & \bf 100 & 99.85 & lao\_Laoo & 9.39 & 94.81 & 96.94 & \bf 100 \\
deu\_Latn & \bf 100 & - & \bf 100 & 99.95 & lij\_Latn & 88.88 & \bf 99.85 & 98.86 & \bf 99.85 \\
dik\_Latn & 21.44 & 74.11 & 57.71 & \bf 82.21 & lim\_Latn & 83.1 & 85.23 & 98.72 & \bf 99.75 \\
dyu\_Latn & 13.39 & \bf 75.89 & 47.73 & 70.06 & lin\_Latn & 34.19 & 99.56 & 72.58 & \bf 99.7 \\
dzo\_Tibt & 0.25 & 92.54 & 92.54 & \bf 98.37 & lit\_Latn & 99.56 & - & \bf 99.6 & 99.46 \\
ell\_Grek & 99.9 & - & \bf 100 & \bf 100 & lmo\_Latn & 78.9 & 98.22 & 97.48 & \bf 99.7 \\
eng\_Latn & - & - & - & - & ltg\_Latn & 78.26 & 99.65 & 95.5 & \bf 99.85 \\
epo\_Latn & \bf 100 & - & \bf 100 & \bf 100 & ltz\_Latn & 66.65 & 99.01 & \bf 100 & 99.95 \\
est\_Latn & 99.85 & - & \bf 100 & 99.85 & lua\_Latn & 34.73 & 96.89 & 70.95 & \bf 97.63 \\
eus\_Latn & 99.8 & - & 99.95 & \bf 100 & lug\_Latn & 22.28 & 97.08 & 80.88 & \bf 98.67 \\
ewe\_Latn & 10.67 & 96.15 & 56.47 & \bf 96.54 & luo\_Latn & 16.21 & 98.76 & 59.19 & \bf 99.6 \\
fao\_Latn & 88.09 & 96.29 & \bf 99.95 & \bf 99.95 & lus\_Latn & 16.7 & 95.06 & 71.29 & \bf 97.97 \\
fij\_Latn & 22.08 & 98.57 & 59.58 & \bf 99.41 & lvs\_Latn & 99.9 & - & \bf 100 & 99.75 \\
fin\_Latn & 99.85 & - & \bf 99.9 & 99.6 & mag\_Deva & 96.1 & 99.46 & \bf 100 & 99.75 \\
fon\_Latn & 10.38 & 81.08 & 47.88 & \bf 84.63 & mai\_Deva & 88.19 & 95.6 & \bf 100 & \bf 100 \\
fra\_Latn & 99.95 & - & \bf 100 & \bf 100 & mal\_Mlym & 99.06 & 99.51 & \bf 99.9 & 99.46 \\
fur\_Latn & 86.17 & 99.9 & 98.96 & \bf 100 & mar\_Deva & 98.91 & 98.52 & \bf 100 & 99.9 \\
fuv\_Latn & 17.14 & 66.06 & 63.14 & \bf 79.35 & min\_Arab & 4.99 & - & 30.63 & \bf 82.46 \\
\hline
\end{tabular}
\caption{The averaged bidirectional similarity search accuracy (\texttt{xx} $\leftrightarrow$ English) on the Flores-200 benchmark  (Part I).}
\label{tab:flores200-en-1}
\end{table*}

\begin{table*}[h]\small
\centering
\begin{tabular}{c | c c c c | c | c c c c} 
Language & LASER2 & LASER3 & LaBSE & MuSR & Language & LASER2 & LASER3 & LaBSE & MuSR \\
\hline
\hline
min\_Latn & 61.46 & 99.56 & 97.13 & \bf 99.9 & spa\_Latn & 99.6 & - & \bf 99.9 & 99.51 \\
mkd\_Cyrl & \bf 100 & - & \bf 100 & 99.95 & srd\_Latn & 89.08 & 99.9 & 99.16 & \bf 100 \\
mlt\_Latn & 25.4 & 99.9 & \bf 100 & \bf 100 & srp\_Cyrl & 99.9 & - & \bf 100 & 99.9 \\
mni\_Beng & 8.4 & 98.27 & 36.81 & \bf 99.26 & ssw\_Latn & 17 & 99.36 & 96.34 & \bf 99.6 \\
mos\_Latn & 17.39 & 81.97 & 54.35 & \bf 86.31 & sun\_Latn & 61.02 & 99.41 & 99.8 & \bf 99.9 \\
mri\_Latn & 18.97 & 97.88 & \bf 99.51 & 99.36 & swe\_Latn & \bf 100 & - & \bf 100 & \bf 100 \\
mya\_Mymr & 83.65 & 98.22 & \bf 99.7 & 99.36 & swh\_Latn & 98.72 & 99.21 & \bf 100 & \bf 100 \\
nld\_Latn & 99.7 & - & \bf 100 & 99.51 & szl\_Latn & 94.86 & \bf 99.21 & 98.86 & \bf 99.21 \\
nno\_Latn & 98.86 & - & \bf 99.9 & \bf 99.9 & tam\_Taml & 82.07 & 99.56 & \bf 100 & 99.41 \\
nob\_Latn & 99.6 & - & \bf 99.9 & 99.75 & taq\_Latn & 38.09 & 72.68 & 55.58 & \bf 76.19 \\
npi\_Deva & 68.63 & 97.63 & \bf 99.7 & 99.41 & taq\_Tfng & 2.08 & - & 16.45 & \bf 61.17 \\
nso\_Latn & 22.73 & 99.7 & 99.06 & \bf 99.9 & tat\_Cyrl & 21 & 95.7 & \bf 100 & 99.8 \\
nus\_Latn & 8.6 & 90.27 & 43.03 & \bf 96.79 & tel\_Telu & 96.54 & 99.01 & \bf 100 & 99.7 \\
nya\_Latn & 31.52 & 99.41 & 99.6 & \bf 99.8 & tgk\_Cyrl & 6.92 & 98.86 & \bf 99.75 & 99.7 \\
oci\_Latn & 99.6 & - & 99.95 & \bf 100 & tgl\_Latn & 90.22 & 99.95 & \bf 100 & \bf 100 \\
ory\_Orya & 3.41 & 99.51 & \bf 100 & 99.46 & tha\_Thai & 99.56 & \bf 99.75 & 94.02 & \bf 99.75 \\
pag\_Latn & 46.84 & 98.52 & 87.85 & \bf 99.16 & tir\_Ethi & 5.53 & \bf 98.72 & 75.94 & 98.52 \\
pan\_Guru & 3.06 & 99.65 & \bf 100 & 99.9 & tpi\_Latn & 30.39 & 99.75 & 83.05 & \bf 100 \\
pap\_Latn & 78.36 & 99.8 & 98.47 & \bf 100 & tsn\_Latn & 17.19 & 98.47 & 97.97 & \bf 98.76 \\
pbt\_Arab & 29.99 & 99.41 & \bf 100 & 99.7 & tso\_Latn & 22.04 & 98.91 & 71.29 & \bf 99.36 \\
pes\_Arab & 98.81 & 98.47 & \bf 100 & 99.75 & tuk\_Latn & 29.94 & 92.54 & \bf 99.95 & 99.75 \\
plt\_Latn & 99.9 & 99.85 & \bf 99.95 & \bf 99.95 & tum\_Latn & 27.12 & 97.78 & 90.46 & \bf 99.06 \\
pol\_Latn & 99.85 & - & \bf 100 & 99.6 & tur\_Latn & 99.06 & 99.16 & \bf 100 & 99.9 \\
por\_Latn & 99.95 & - & \bf 100 & \bf 100 & twi\_Latn & 25.44 & 98.96 & 71.79 & \bf 99.06 \\
prs\_Arab & 98.12 & 97.48 & \bf 100 & 99.75 & tzm\_Tfng & 1.73 & 95.45 & 16.3 & \bf 97.38 \\
quy\_Latn & 19.76 & 71.79 & 57.71 & \bf 93.63 & uig\_Arab & 17.14 & 91.75 & \bf 99.8 & 99.51 \\
ron\_Latn & 99.95 & - & \bf 100 & \bf 100 & ukr\_Cyrl & 99.95 & - & \bf 100 & 99.95 \\
run\_Latn & 19.12 & 99.26 & \bf 99.51 & 99.46 & umb\_Latn & 19.96 & 83.79 & 58.2 & \bf 87.15 \\
rus\_Cyrl & 99.85 & - & \bf 100 & 99.95 & urd\_Arab & 89.28 & 99.46 & \bf 99.9 & 99.56 \\
sag\_Latn & 25.2 & 89.33 & 62.7 & \bf 94.86 & uzn\_Latn & 19.12 & 99.6 & \bf 99.9 & 99.51 \\
san\_Deva & 49.65 & 83.4 & 96.44 & \bf 98.57 & vec\_Latn & 94.32 & 97.18 & 99.8 & \bf 99.95 \\
sat\_Olck & 0.3 & - & 4.15 & \bf 95.41 & vie\_Latn & 99.9 & - & \bf 100 & 99.9 \\
scn\_Latn & 76.63 & 99.26 & 98.42 & \bf 99.85 & war\_Latn & 55.43 & 99.9 & 99.95 & \bf 100 \\
shn\_Mymr & 16.25 & 98.52 & 48.37 & \bf 99.51 & wol\_Latn & 25 & 89.77 & 68.48 & \bf 95.7 \\
sin\_Sinh & 99.65 & 99.16 & \bf 100 & 99.26 & xho\_Latn & 18.33 & \bf 99.8 & 99.7 & \bf 99.8 \\
slk\_Latn & 99.85 & - & \bf 100 & 99.75 & ydd\_Hebr & 11.91 & 95.41 & 99.95 & \bf 100 \\
slv\_Latn & 99.85 & - & \bf 100 & 99.8 & yor\_Latn & 21.25 & 95.06 & \bf 97.43 & 97.18 \\
smo\_Latn & 18.82 & 99.7 & 99.56 & \bf 99.85 & yue\_Hant & 93.53 & - & \bf 100 & 99.85 \\
sna\_Latn & 19.52 & 99.46 & 99.26 & \bf 99.65 & zho\_Hans & 99.56 & - & \bf 100 & 99.6 \\
snd\_Arab & 24.51 & 97.58 & \bf 100 & 99.7 & zho\_Hant & 94.02 & - & \bf 99.95 & 99.46 \\
som\_Latn & 8.55 & 98.07 & 99.65 & \bf 99.7 & zsm\_Latn & 99.11 & 99.9 & \bf 100 & \bf 100 \\
sot\_Latn & 20.85 & 99.8 & 99.9 & \bf 100 & zul\_Latn & 13.19 & 99.85 & 99.85 & \bf 99.9 \\
\hline
\end{tabular}
\caption{The averaged bidirectional similarity search accuracy (\texttt{xx} $\leftrightarrow$ English) on the Flores-200 benchmark  (Part II).}
\label{tab:flores200-en-2}
\end{table*}

\begin{table*}[h]\small
\centering
\begin{tabular}{c | c c c c | c | c c c c} 
Language & LASER2 & LASER3 & LaBSE & MuSR & Language & LASER2 & LASER3 & LaBSE & MuSR \\
\hline
\hline
ace\_Arab & 6.27 & - & 29.2 & \bf 73.57 & gaz\_Latn & 7.51 & 92.59 & 40.96 & \bf 97.88 \\
ace\_Latn & 29.69 & 91.4 & 81.92 & \bf 97.68 & gla\_Latn & 4.84 & 81.27 & \bf 99.85 & 98.76 \\
acm\_Arab & 98.52 & - & \bf 99.9 & 99.56 & gle\_Latn & 5.09 & 92.64 & \bf 99.95 & 98.86 \\
acq\_Arab & 98.76 & - & \bf 99.95 & 99.7 & glg\_Latn & 99.56 & - & \bf 100 & 99.65 \\
aeb\_Arab & 96.99 & - & \bf 98.86 & \bf 98.86 & grn\_Latn & 26.53 & 96.25 & 71.49 & \bf 97.38 \\
afr\_Latn & 97.83 & - & \bf 100 & 99.46 & guj\_Gujr & 2.57 & 98.81 & \bf 100 & 99.65 \\
ajp\_Arab & 98.52 & - & \bf 99.75 & 99.56 & hat\_Latn & 24.31 & 96.25 & \bf 99.21 & 98.42 \\
aka\_Latn & 16.55 & 94.52 & 58.89 & \bf 96.59 & hau\_Latn & 16.11 & 97.13 & \bf 99.11 & 99.01 \\
als\_Latn & 98.91 & - & \bf 100 & 99.21 & heb\_Hebr & 99.21 & - & \bf 100 & 99.51 \\
amh\_Ethi & 47.48 & 99.01 & \bf 99.9 & 99.65 & hin\_Deva & 97.83 & 99.51 & \bf 99.95 & 99.6 \\
apc\_Arab & 98.47 & - & \bf 99.7 & \bf 99.7 & hne\_Deva & 87.15 & 96.64 & 98.91 & \bf 99.11 \\
arb\_Arab & 99.56 & - & \bf 100 & 99.7 & hrv\_Latn & 99.31 & - & \bf 99.95 & 99.56 \\
arb\_Latn & 5.78 & - & \bf 36.51 & 31.72 & hun\_Latn & 99.51 & - & \bf 100 & 99.8 \\
ars\_Arab & 99.51 & - & \bf 100 & 99.6 & hye\_Armn & 77.72 & 98.52 & \bf 100 & 99.6 \\
ary\_Arab & 87.5 & - & 96.1 & \bf 97.48 & ibo\_Latn & 13.29 & 96.94 & \bf 99.01 & 98.42 \\
arz\_Arab & 98.17 & - & \bf 99.7 & 99.31 & ilo\_Latn & 30.93 & 99.16 & 81.82 & \bf 99.41 \\
asm\_Beng & 49.31 & 91.5 & \bf 99.51 & 99.11 & ind\_Latn & 98.22 & 99.31 & \bf 100 & 99.65 \\
ast\_Latn & 95.06 & - & \bf 99.75 & 98.91 & isl\_Latn & 97.48 & - & \bf 99.85 & 99.11 \\
awa\_Deva & 93.97 & 91.9 & \bf 99.06 & 98.86 & ita\_Latn & 99.65 & - & \bf 100 & 99.8 \\
ayr\_Latn & 11.26 & 75.59 & 46.25 & \bf 92.54 & jav\_Latn & 45.36 & 98.02 & \bf 100 & 99.46 \\
azb\_Arab & 41.01 & 55.34 & 81.57 & \bf 92.59 & jpn\_Jpan & 99.21 & - & \bf 100 & 99.41 \\
azj\_Latn & 49.06 & 97.78 & \bf 99.6 & 98.57 & kab\_Latn & 70.75 & 89.97 & 37.2 & \bf 95.8 \\
bak\_Cyrl & 12.35 & 96.15 & 84.73 & \bf 99.56 & kac\_Latn & 10.03 & 86.51 & 48.62 & \bf 95.9 \\
bam\_Latn & 13.24 & 87.25 & 48.27 & \bf 92 & kam\_Latn & 21.74 & 72.92 & 58.79 & \bf 79.79 \\
ban\_Latn & 46.25 & 97.48 & 95.9 & \bf 98.42 & kan\_Knda & 1.88 & 97.53 & \bf 100 & 99.46 \\
bel\_Cyrl & 67.98 & 97.53 & \bf 100 & 98.62 & kas\_Arab & 31.42 & 97.08 & 86.46 & \bf 98.17 \\
bem\_Latn & 24.85 & 96.99 & 72.92 & \bf 97.78 & kas\_Deva & 25.84 & 89.67 & 72.38 & \bf 92.93 \\
ben\_Beng & 99.21 & 97.38 & \bf 99.95 & 99.6 & kat\_Geor & 70.01 & 94.91 & \bf 100 & 99.06 \\
bho\_Deva & 82.02 & 96.25 & 98.72 & \bf 99.36 & kaz\_Cyrl & 47.08 & 97.33 & \bf 99.8 & 99.31 \\
bjn\_Arab & 6.08 & - & 24.26 & \bf 74.7 & kbp\_Latn & 9.88 & 83.35 & 45.31 & \bf 90.91 \\
bjn\_Latn & 69.12 & 98.22 & 96.64 & \bf 98.81 & kea\_Latn & 64.62 & 92.69 & 93.53 & \bf 99.21 \\
bod\_Tibt & 2.42 & 76.33 & \bf 98.07 & 96.84 & khk\_Cyrl & 11.46 & 95.95 & \bf 100 & 99.46 \\
bos\_Latn & 99.7 & - & \bf 100 & 99.51 & khm\_Khmr & 69.07 & 88.24 & 97.83 & \bf 99.31 \\
bug\_Latn & 26.38 & 92.34 & 76.53 & \bf 94.96 & kik\_Latn & 8.05 & 95.36 & 57.56 & \bf 96.54 \\
bul\_Cyrl & 99.36 & - & \bf 100 & 99.6 & kin\_Latn & 15.02 & 98.32 & \bf 99.56 & 99.21 \\
cat\_Latn & 99.51 & - & \bf 100 & 99.51 & kir\_Cyrl & 26.73 & 93.82 & \bf 99.8 & 98.96 \\
ceb\_Latn & 46.74 & 98.52 & \bf 99.95 & 99.56 & kmb\_Latn & 20.8 & 80.29 & 51.43 & \bf 84.78 \\
ces\_Latn & 99.6 & - & \bf 100 & 99.8 & kmr\_Latn & 14.87 & 92.98 & \bf 99.65 & 98.96 \\
cjk\_Latn & 21.15 & 62.06 & 53.26 & \bf 73.22 & knc\_Arab & 7.41 & \bf 29.74 & 20.11 & 17.59 \\
ckb\_Arab & 3.51 & 98.86 & 37.35 & \bf 99.16 & knc\_Latn & 12.75 & 83.3 & 50.49 & \bf 88.29 \\
crh\_Latn & 71.25 & 98.57 & 99.21 & \bf 99.56 & kon\_Latn & 31.82 & 94.86 & 61.71 & \bf 98.07 \\
cym\_Latn & 12.99 & 96.15 & \bf 100 & 99.65 & kor\_Hang & 98.67 & - & \bf 99.9 & 99.65 \\
dan\_Latn & 99.56 & - & \bf 100 & 99.46 & lao\_Laoo & 7.81 & 88.59 & 96.59 & \bf 99.6 \\
deu\_Latn & 99.6 & - & \bf 100 & 99.7 & lij\_Latn & 73.96 & 98.62 & 95.45 & \bf 99.41 \\
dik\_Latn & 15.22 & 61.91 & 50.15 & \bf 73.07 & lim\_Latn & 70.06 & 71.1 & 96.59 & \bf 98.52 \\
dyu\_Latn & 9.83 & \bf 65.51 & 41.21 & 62.3 & lin\_Latn & 28.61 & 97.68 & 61.81 & \bf 98.47 \\
dzo\_Tibt & 0.3 & 88.54 & 89.03 & \bf 97.08 & lit\_Latn & 99.21 & - & \bf 99.51 & 99.26 \\
ell\_Grek & 99.36 & - & \bf 100 & 99.7 & lmo\_Latn & 60.67 & 93.28 & 92.59 & \bf 97.92 \\
eng\_Latn & 99.56 & - & \bf 100 & 99.6 & ltg\_Latn & 66.35 & 98.67 & 91.35 & \bf 99.11 \\
epo\_Latn & 99.26 & - & \bf 100 & 99.56 & ltz\_Latn & 51.14 & 94.37 & \bf 99.85 & 99.7 \\
est\_Latn & 99.41 & - & \bf 99.95 & 99.8 & lua\_Latn & 26.88 & 90.46 & 62.06 & \bf 93.58 \\
eus\_Latn & 98.12 & - & \bf 99.95 & 99.7 & lug\_Latn & 15.51 & 92.05 & 69.52 & \bf 96.1 \\
ewe\_Latn & 8.2 & 93.28 & 50.59 & \bf 94.71 & luo\_Latn & 11.76 & 94.47 & 51.04 & \bf 97.68 \\
fao\_Latn & 76.53 & 87.9 & \bf 99.75 & 99.41 & lus\_Latn & 12.8 & 88.44 & 63.64 & \bf 95.85 \\
fij\_Latn & 15.56 & 96.15 & 51.09 & \bf 97.78 & lvs\_Latn & 99.51 & - & \bf 99.95 & 99.6 \\
fin\_Latn & 99.36 & - & \bf 99.85 & 99.51 & mag\_Deva & 91.9 & 98.62 & 99.65 & \bf 99.75 \\
fon\_Latn & 8 & 73.12 & 43.38 & \bf 79.2 & mai\_Deva & 81.92 & 90.46 & 99.65 & \bf 99.8 \\
fra\_Latn & 99.6 & - & \bf 100 & 99.65 & mal\_Mlym & 97.04 & 98.76 & \bf 99.85 & 99.21 \\
fur\_Latn & 72.83 & 98.37 & 96.39 & \bf 99.51 & mar\_Deva & 96.29 & 96.39 & \bf 99.9 & 99.6 \\
fuv\_Latn & 11.91 & 55.58 & 55.88 & \bf 71.15 & min\_Arab & 3.75 & - & 23.67 & \bf 72.78 \\
\hline
\end{tabular}
\caption{The averaged bidirectional similarity search accuracy (\texttt{xx} $\leftrightarrow$ Chinese) on the Flores-200 benchmark  (Part I).}
\label{tab:flores200-zh-1}
\end{table*}

\begin{table*}[h]\small
\centering
\begin{tabular}{c | c c c c | c | c c c c} 
Language & LASER2 & LASER3 & LaBSE & MuSR & Language & LASER2 & LASER3 & LaBSE & MuSR \\
\hline
\hline
min\_Latn & 50.89 & 97.88 & 93.97 & \bf 99.31 & spa\_Latn & 99.36 & - & \bf 99.9 & 99.11 \\
mkd\_Cyrl & 99.6 & - & \bf 100 & 99.85 & srd\_Latn & 72.48 & 96.34 & 96.54 & \bf 99.21 \\
mlt\_Latn & 18.92 & 98.72 & \bf 100 & 99.56 & srp\_Cyrl & 98.62 & - & \bf 100 & 99.7 \\
mni\_Beng & 7.36 & 93.82 & 30.63 & \bf 98.52 & ssw\_Latn & 11.86 & 98.22 & 90.56 & \bf 98.57 \\
mos\_Latn & 13.39 & 72.73 & 47.68 & \bf 79.79 & sun\_Latn & 51.28 & 97.48 & \bf 99.7 & 99.11 \\
mri\_Latn & 14.72 & 94.27 & \bf 98.42 & 97.58 & swe\_Latn & 99.65 & - & \bf 100 & 99.6 \\
mya\_Mymr & 79 & 96.64 & \bf 99.65 & 99.26 & swh\_Latn & 95.95 & 96.05 & \bf 99.95 & 99.21 \\
nld\_Latn & 98.96 & - & \bf 100 & 99.46 & szl\_Latn & 85.08 & 98.27 & 97.83 & \bf 98.62 \\
nno\_Latn & 95.06 & - & \bf 99.85 & 99.41 & tam\_Taml & 76.28 & 98.02 & \bf 99.95 & 98.76 \\
nob\_Latn & 98.27 & - & \bf 99.8 & 99.41 & taq\_Latn & 27.72 & 59.88 & 49.16 & \bf 69.17 \\
npi\_Deva & 61.56 & 94.27 & \bf 99.7 & 99.06 & taq\_Tfng & 1.68 & - & 13.69 & \bf 53.85 \\
nso\_Latn & 17.34 & 98.52 & 96.1 & \bf 99.01 & tat\_Cyrl & 16.85 & 91.6 & \bf 100 & 99.6 \\
nus\_Latn & 7.36 & 79.5 & 36.26 & \bf 92.34 & tel\_Telu & 90.81 & 97.58 & \bf 100 & 99.31 \\
nya\_Latn & 24.56 & 97.78 & \bf 98.81 & 98.72 & tgk\_Cyrl & 4.79 & 96.89 & \bf 99.75 & 99.16 \\
oci\_Latn & 95.6 & - & \bf 99.7 & 99.56 & tgl\_Latn & 77.37 & 99.31 & \bf 99.9 & 99.51 \\
ory\_Orya & 2.77 & 99.01 & \bf 100 & 99.31 & tha\_Thai & \bf 99.36 & 99.21 & 93.73 & 99.31 \\
pag\_Latn & 35.67 & 96.1 & 82.91 & \bf 97.88 & tir\_Ethi & 5.93 & 95.75 & 68.73 & \bf 97.63 \\
pan\_Guru & 2.57 & 98.57 & \bf 100 & 99.51 & tpi\_Latn & 22.83 & 94.52 & 73.57 & \bf 99.06 \\
pap\_Latn & 63.34 & 98.96 & 95.36 & \bf 99.65 & tsn\_Latn & 12.9 & 96.94 & 94.81 & \bf 97.53 \\
pbt\_Arab & 26.73 & 97.33 & \bf 99.36 & 99.31 & tso\_Latn & 16.7 & 97.68 & 59.14 & \bf 98.57 \\
pes\_Arab & 97.92 & 95.85 & \bf 100 & 99.6 & tuk\_Latn & 26.58 & 85.72 & \bf 99.75 & 99.41 \\
plt\_Latn & 99.41 & 98.86 & \bf 99.56 & 99.06 & tum\_Latn & 21.49 & 95.5 & 85.47 & \bf 97.53 \\
pol\_Latn & 99.26 & - & \bf 99.95 & 99.6 & tur\_Latn & 98.12 & 97.73 & \bf 100 & 99.75 \\
por\_Latn & 99.56 & - & \bf 100 & 99.51 & twi\_Latn & 17.64 & 95.55 & 62.01 & \bf 97.08 \\
prs\_Arab & 97.28 & 93.92 & \bf 100 & 99.7 & tzm\_Tfng & 1.63 & 87.65 & 14.33 & \bf 92.49 \\
quy\_Latn & 14.48 & 61.76 & 51.78 & \bf 88.64 & uig\_Arab & 14.08 & 86.71 & \bf 99.85 & 99.21 \\
ron\_Latn & 99.06 & - & \bf 100 & 99.56 & ukr\_Cyrl & 99.26 & - & \bf 100 & 99.65 \\
run\_Latn & 14.97 & 97.68 & 98.12 & \bf 98.86 & umb\_Latn & 15.66 & 75.59 & 51.73 & \bf 79.35 \\
rus\_Cyrl & 98.96 & - & \bf 100 & 99.75 & urd\_Arab & 86.12 & 98.37 & \bf 99.8 & 99.46 \\
sag\_Latn & 19.61 & 80.78 & 54.5 & \bf 89.58 & uzn\_Latn & 15.61 & 98.27 & \bf 99.85 & 99.21 \\
san\_Deva & 43.63 & 78.26 & 93.08 & \bf 97.48 & vec\_Latn & 85.42 & 89.87 & 98.47 & \bf 99.51 \\
sat\_Olck & 0.25 & - & 2.62 & \bf 91.25 & vie\_Latn & 99.41 & - & \bf 100 & 99.51 \\
scn\_Latn & 61.61 & 97.04 & 95.16 & \bf 98.86 & war\_Latn & 39.72 & 99.16 & \bf 99.56 & 99.41 \\
shn\_Mymr & 12.5 & 95.11 & 42.29 & \bf 98.67 & wol\_Latn & 18.48 & 76.93 & 60.77 & \bf 90.46 \\
sin\_Sinh & 98.47 & 97.92 & \bf 99.9 & 99.06 & xho\_Latn & 12.3 & 98.76 & 98.91 & \bf 99.11 \\
slk\_Latn & 99.41 & - & \bf 100 & 99.56 & ydd\_Hebr & 9.63 & 77.77 & \bf 99.36 & 99.01 \\
slv\_Latn & 99.26 & - & \bf 100 & 99.41 & yor\_Latn & 15.07 & 90.76 & 93.73 & \bf 94.32 \\
smo\_Latn & 13.44 & 98.52 & \bf 99.06 & 98.62 & yue\_Hant & 93.68 & - & \bf 100 & 99.85 \\
sna\_Latn & 13.93 & 97.58 & 97.68 & \bf 98.72 & zho\_Hans & - & - & - & - \\
snd\_Arab & 21.34 & 94.07 & \bf 99.7 & 99.06 & zho\_Hant & 94.32 & - & \bf 99.9 & 99.56 \\
som\_Latn & 6.97 & 93.68 & 98.67 & \bf 98.76 & zsm\_Latn & 98.42 & 99.46 & \bf 100 & 99.51 \\
sot\_Latn & 14.48 & 99.11 & 98.76 & \bf 99.16 & zul\_Latn & 9.29 & 99.26 & \bf 99.51 & 99.31 \\
\hline
\end{tabular}
\caption{The averaged bidirectional similarity search accuracy (\texttt{xx} $\leftrightarrow$ Chinese) on the Flores-200 benchmark  (Part II).}
\label{tab:flores200-zh-2}
\end{table*}

\end{document}